# Interpretable machine learning optimization (InterOpt) for operational parameters: a case study of highly-efficient shale gas development

Yuntian Chen [1], Dongxiao Zhang [2,3,*], Qun Zhao [4], and Dexun Liu [4]

1. *Eastern Institute for Advanced Study, Yongriver Institute of Technology, Zhejiang, P. R. China*
2. *Department of Mathematics and Theories, Peng Cheng Laboratory, Guangdong, P. R. China*
3. *National Center for Applied Mathematics Shenzhen (NCAMS), Southern University of Science and Technology, Guangdong, P. R. China*
4. *Research Institute of Petroleum Exploration and Development, CNPC, Beijing, P. R. China*

[*] Corresponding author

**Abstract:** An algorithm named InterOpt for optimizing operational parameters is proposed based on interpretable machine learning, and is demonstrated via optimization of shale gas development. InterOpt consists of three parts: a neural network is used to construct an emulator of the actual drilling and hydraulic fracturing process in the vector space (i.e., virtual environment); the Sharpley value method in interpretable machine learning is applied to analyzing the impact of geological and operational parameters in each well (i.e., single well feature impact analysis); and ensemble randomized maximum likelihood (EnRML) is conducted to optimize the operational parameters to comprehensively improve the efficiency of shale gas development and reduce the average cost. In the experiment, InterOpt provides different drilling and fracturing plans for each well according to its specific geological conditions, and finally achieved an average cost reduction of 9.7% for a case study with 104 wells.

## 1. Introduction

In recent years, the issue of highly-efficient shale gas development has received widespread attention. Petroleum engineers usually design drilling and fracturing plans based on experience or design of experiments (DOE) in practice. DOE is a method to study the relationship between multiple variables via experiments. Although this practice-based method has improved the recovery rate and economic benefits to a certain extent, the cost of iterative optimization remains high. In addition, the inability to perform multiple drilling and fracturing experiments with different operational parameters at the same location further limits the application of DOE.

In order to improve the economic benefits of shale gas development, a feasible method to optimize the operational parameters is to: (1) build a virtual environment that reflects real-world scenarios; (2) analyze the impact of different operational parameters based on the environment; and (3) optimize the drilling and fracturing plans of different wells in consideration of specific geological conditions. However, many challenges exist in the above three steps/tasks, as shown in Table 1.

**Table 1.** Challenges of conventional methods to optimize operational parameters.

| Tasks | Conventional methods | Challenges |
|---|---|---|
| Virtual environment construction | Numerical simulations (Cipolla et al., 2009; Sun et al., 2015; Yao et al., 2013) | High computational cost |
| Feature impact analysis | Global interpretation methods (Guo et al., 2017; Li et al., 2021; Luo et al., 2019; Wang et al., 2019) | Cannot evaluate each well according to specific conditions |
| | | Cannot consider the interaction between multiple features |
| | Cross-validation based methods (Wang and Chen, 2019) | High computational cost |
| Operational plan optimization | Visualization (Wang and Chen, 2019) and trend graph (Luo et al., 2019) | Limited to low dimensions |
| | Genetic algorithms (Zhao et al., 2020) and digital twin-based design of experiments (Shen et al., 2021) | Low optimization efficiency |

In terms of constructing a virtual environment that reflects real-world scenarios for optimization algorithms to run on, the conventional method is to simulate the underground physical field and the drilling and fracturing process (Cipolla et al., 2009; Sun et al., 2015; Yao et al., 2013). However, the computational cost of numerical simulation is high, and optimization of parameters based on simulations is extremely time-consuming.

In terms of feature impact analysis, researchers have utilized many methods to determine the most important features (i.e., main controlling factors), such as sensitivity analysis (Wang et al., 2019), correlation analysis (Guo et al., 2017; Luo et al., 2019), and intersection diagram (Li et al., 2021). Although these methods are simple and easy to implement, they can only provide the global main controlling factors and cannot analyze the influence of different features in a specific single well (i.e., local interpretation). However, the main controlling factors of different wells may vary greatly in practice, and the shortcomings of each well are not identical, which indicates that different wells require different optimization plans.

In addition, it is difficult for existing methods based on sensitivity or correlation to consider the interaction between multiple features. For example, if the correlation between the amount of quartz sand used in fracturing and the production is counted, a negative correlation is usually found. This leads to the incorrect conclusion that quartz sand is not helpful to increasing production. This misleading phenomenon is resultant from the fact that petroleum engineers will adopt different drilling and fracturing plans for wells with different geological conditions. Indeed, engineers tend to use more quartz sand to increase expected production for wells with poor geological conditions (but the final production is still low). Moreover, they reduce the quartz sand in consideration of cost for the wells that are expected to be highly productive. Therefore, to objectively evaluate the impact of each feature, compared with traditional correlation or sensitivity methods, the interaction between all features (e.g., geological conditions) should be considered comprehensively. Cross-validation (e.g., RFECV (Wang and Chen, 2019)) is an attempt to comprehensively consider the interaction, but it is essentially a traversal algorithm and is extremely computationally-expensive. For instance, RFECV repeatedly trains 110 different neural networks to analyze 11 features in the experiment (Wang and Chen, 2019). Consequently, an urgent need exists to find a feature impact analysis method that can not only determine the shortcomings of each well, but also comprehensively consider the interaction between all features.

In terms of operational parameters optimization, extant literature is mainly based on visualization (Wang and Chen, 2019) or trend graph (Luo et al., 2019) to identify the trend of target features under different combinations of main controlling factors, and to find the optimal combination. Due to the limitation of the drawing dimension, the above methods can only optimize no more than three factors at the same time, and their essence is to traverse all possible combinations of the selected main controlling factors, which is inefficient. Recently, some studies have used genetic algorithms (Zhao et al., 2020) and digital twin technology (Shen et al., 2021) to optimize production. Although the genetic algorithm can optimize multiple features simultaneously, it cannot effectively determine the optimization direction, which results in low optimization efficiency. Digital twin-based optimization is essentially a DOE for the digital world, and does not take advantage of efficient optimization algorithms in machine learning. The challenge in this step is to optimize all adjustable operational parameters simultaneously, and to improve the optimization strategy to avoid traversing the solution space.

Considering the aforementioned problems, *a salient question* is: are there any optimization methods that are able to customize operational parameters of each well according to their specific geological conditions (local interpretation), and comprehensively consider the interaction between all features (including geological and operational parameters), but do not rely on the computationally-expensive traversal algorithm?

To answer this question, this study proposes the interpretable machine learning optimization (InterOpt) algorithm to customize optimal operational parameters at the single well level. The major characteristics and advantages of InterOpt include the following:

- In order to reduce the computational cost of simulation, a neural network-based emulator is constructed as the surrogate model in InterOpt. Neural networks can effectively find complex mapping relationships among variables, which have become a powerful tool for establishing surrogate models and are widely used in petroleum engineering, such as sweet point searching

- (Tang et al., 2021), production prediction (Liu et al., 2020; Song et al., 2020), lithology identification (Rogers et al., 1992), and well log analysis (Luo et al., 2022; Zhang et al., 2018), etc. Meanwhile, a large amount of data has been generated and stored in shale gas development, which forms a solid foundation for training data-driven algorithms. Therefore, a neural network-based emulator is able to mimic or imitate real-time behaviors of shale gas development with low cost and high speed.
- Interpretable machine learning is utilized to analyze the specific impact of all features in each well, which is also the basis for InterOpt to find the operational parameters to be optimized (i.e., the shortcomings of each well). Interpretable machine learning includes model-based methods and model-agnostic methods (Molnar, 2020). Model-based methods often adopt models with interpretable ability (e.g., decision tree) (Huang et al., 2007). Although these methods are effective, they are only applicable to certain specific machine learning models. Model-agnostic methods have no requirements for the model to be explained, which offers high flexibility. In addition, there are global interpretation (Alvarez Melis and Jaakkola, 2018; Chen et al., 2018; Shwartz-Ziv and Tishby, 2017; Zhang et al., 2019) and local interpretation (Kim et al., 2018; Liang et al., 2021; Ribeiro et al., 2016), as well. The global methods are similar to sensitivity or correlation analysis, and are based on an overall understanding of features, which is not suitable for single well optimization. In contrast, local methods can evaluate the specific contribution of a feature under different feature combinations (i.e., the interpretability for each sample). InterOpt adopts a model-agnostic local interpretation method based on Shapley value (Aumann and Hart, 2002; Lundberg and Lee, 2017) to analyze the impact of geological and operational parameters on the average cost of shale gas wells. It is worth mentioning that the Shapley value method evaluates the impact of different features from local to global (i.e., from the single well to the whole block). Therefore, it can not only evaluate the feature impacts in the overall sample, but it can also determine the shortcomings of each well and optimize the operational parameters. Furthermore, since the Shapley value method uses the emulator based on the neural network for analysis, it contains all variables and is capable to comprehensively consider the interaction between all features, which is another advantage over conventional methods.
- The solution space is huge since all adjustable features (i.e., operational parameters) are simultaneously optimized, and it is unrealistic to find the optimal parameters by traversal algorithms (e.g., visualization methods, trend graph methods, and cross-validation methods). In order to reduce the computational cost of the optimization process, InterOpt utilizes covariance-based ensemble randomized maximum likelihood (EnRML) for optimization, in which the optimal parameters are determined by iterative optimization without traversing the whole solution space.

InterOpt can not only comprehensively consider the interaction between features, but it can also evaluate the feature impact of single wells and optimize the operational parameters with low computational cost, which is a feasible way to improve the economic benefits of shale gas development. This paper contains five sections. Section 2 introduces the details of InterOpt. Section 2.1 proposes a neural network-based emulator. Section 2.2 presents the interpretable machine learning method used to analyze the feature impact. In Section 2.3, InterOpt uses the EnRML to optimize the operational parameters according to the specific condition of each well. Section 3 includes the experiment results. Finally, the conclusion and discussion are given in section 4 and section 5, respectively.

## 2. Methodology

InterOpt includes three parts: (1) a neural network-based emulator that models the mapping relationship between geological parameters, operational parameters, and targets (e.g., average cost); (2) feature impact analysis that uses interpretable machine learning to determine the main controlling factors for each well; and (3) operational parameters optimization that obtains the optimal drilling and fracturing plan according to the specific geological conditions of each single well. The three core steps of InterOpt are shown in Fig. 1.

### 2.1 Neural network-based emulator

In order to project the real-world physical quantities into vector space, an emulator is constructed to mimic the mapping relationship between different features in shale gas development. Fig. 2 illustrates the process of transforming real-world problems into vector space. InterOpt uses a neural network to build the emulator of the real world, as shown in Eq. 1:

$$y = f(x) = \sigma(w^m \cdots \sigma(w^3 \sigma(w^2 \sigma(w^1 x + b^1) + b^2) + b^3) \cdots + b^m) \tag{1}$$

where $x$ represents the inputs (i.e., geological parameters and operational parameters); $y$ denotes the outputs/targets (e.g., average cost); $\sigma$ is the activation function; and $w^m$ and $b^m$ are the weights and bias of the $m$th layer, respectively.

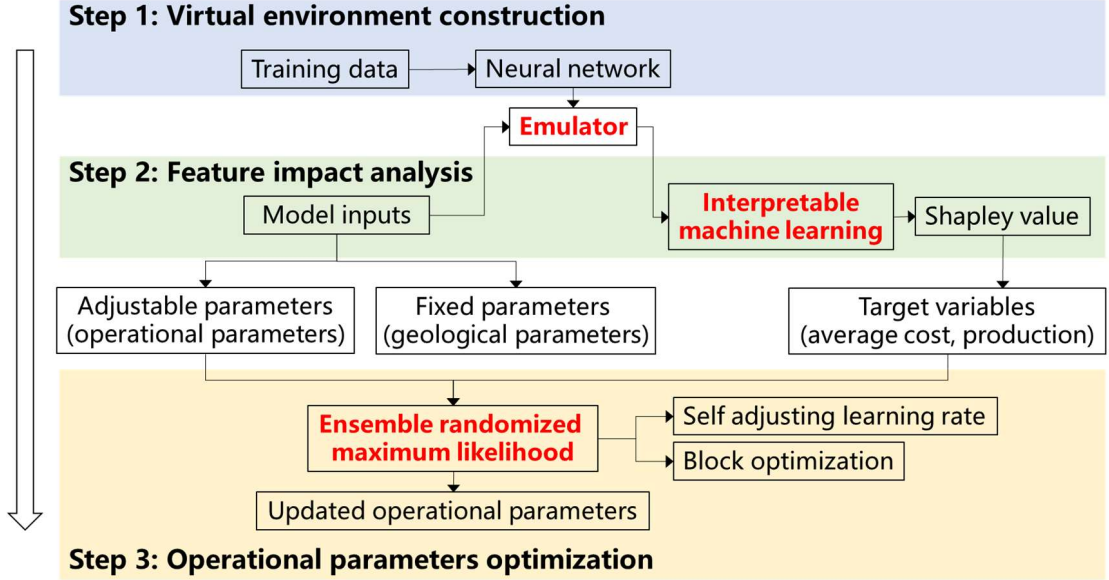

**Fig. 1.** Core steps of interpretable machine learning optimization (InterOpt).

The neural network has a large number of undetermined parameters, which makes the emulator have strong fitting and expression ability (Cybenko, 1989; Hornik, 1991), and can describe the complex nonlinear mapping relationships between different features. The emulator is the basis for subsequent feature impact analysis and operational parameters optimization.

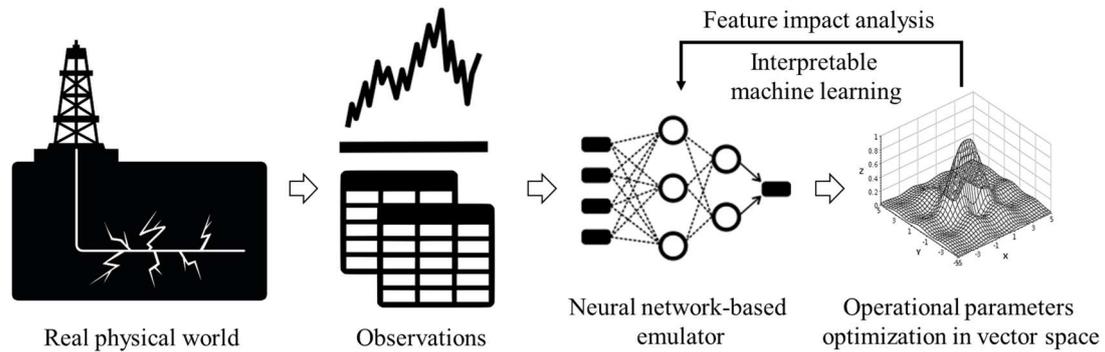

**Fig. 2.** Illustration of the relationship between the real world and the emulator in InterOpt.

### 2.2 Interpretable machine learning and feature impact analysis

InterOpt adopts an interpretable machine learning algorithm, called the Shapley value method (Aumann and Hart, 2002; Lundberg and Lee, 2017), to evaluate the contribution of various geological and operational parameters on the targets (e.g., average cost), and identify the main controlling factors of each well in the drilling and fracturing process. The Shapley value method is a model-agnostic local interpretation method, which can analyze the black box model, and is especially suitable for the neural network-based emulator in InterOpt. The Shapley value comes from the concept of cooperative game theory, which aims to fairly evaluate the contribution of each participant in the game process (Roth, 1988; Shapley, 1951). In terms of machine learning, if a specific regression or classification mission is regarded as a game, and each feature in the model is regarded as a player participating in the game, the theoretical model in cooperative game theory can be applied to machine learning to explain the impact of different features (Molnar, 2020).

Specifically, InterOpt regards the emulator in section 2.1 as a game process. Running each data sample once in the emulator is equivalent to all features participating in a game. The following example in cooperative game theory is helpful to intuitively understand the Shapley value: suppose that a group of players (i.e., operational parameters) randomly participate in a game (i.e., shale gas well development under given geological condition), and the Shapley value of the player (i.e., the contribution of the feature) is the change in the gain of the game (e.g., average cost) after the player joins the game. By calculating the contribution of each player (i.e., feature) in a specific game (i.e., drilling and fracturing process), we can obtain the contribution of each feature in this specific data sample, which is the local interpretation. In addition, by averaging the contribution of a feature in all samples, the overall importance of a feature can be obtained, which is the global interpretation.

It should be noted that before the player enters the game, the state of the game can be different, and thus the change is the average of all possible scenarios. In other words, the contribution of a player (i.e., feature) is defined as the weighted average (i.e., mathematical expectation) of value increment when the player joins the game in all possible scenarios. Therefore, from a mathematical point of view, the Shapley value is defined as the average expected marginal contribution of a feature in all possible scenarios.

According to the aforementioned definition, it is necessary to construct a value function $val_x(S)$ to calculate the Shapley value. The value function describes the marginal contribution of the features contained in the subset $S$ to the features not contained in the subset $S$. In other words, $val_x(S)$ evaluates the difference of emulator output (i.e., gain of the game) between the scenario with the current value of the features in the subset $S$ and the scenario with a random value of all features, as shown in Eq. 2:

$$val(S) = E_{C_U S}(f(X)) - E_X(f(X))$$
$$= \int f(x_1, ..., x_n) d\mathrm{P}_{C_U S} - E_X(f(X)) \tag{2}$$

where $val_x(S)$ denotes the value function; $U$ is the complete set of all $n$ input features; $S$ is a subset of the feature set $U$, representing one of all possible feature combinations; $C_U S$ is the complement of $S$; $f()$ represents the mapping relationship between input features and output features (i.e., the emulator in section 2.1); $E_{C_U S}(f(X))$ represents the mathematical expectation of the gain of the game when all features not included in the subset $S$ take different values; and $E_X(f(X))$ represents the mathematical expectation of the gain of the game when all features take different values.

In order to intuitively demonstrate the meaning of value function $val_x(S)$, the calculation process of $val_x(S)$ when $S$ is a set containing all features, except $X_i$ and $X_j$ (i.e., $S=C_U\{X_i, X_j\}$), is taken as an example in Eq. 3:

$$val(C_U\{X_i, X_j\}) = val(\{x_1, ..., x_{i-1}, x_{i+1}, ..., x_{j-1}, x_{j+1}, ..., x_n\})$$
$$= \iint f(x_1, ..., x_{i-1}, X_i, x_{i+1}, ..., x_{j-1}, X_j, x_{j+1}, ..., x_n) d\mathrm{P}_{X_i, X_j} - E_X(f(X)) \tag{3}$$

where $\{x_1, ..., x_{i-1}, x_{i+1}, ..., x_{j-1}, x_{j+1}, ..., x_n\}$ represents the specific values of the features in the subset $S$; and $X_i$ and $X_j$ are the two features not included in the subset $S$. When calculating the value function $val_x(S)$, the values of these two missing features are processed by integration to obtain the mathematical expectation.

When calculating the Shapley value of a feature, the features contained in subset $S$ have a variety of possible combinations. The feature contribution under different feature combinations is comprehensively considered by weighted summation of value functions in different scenarios. In order to determine the weights of the weighted summation, the number of permutations of each feature combination needs to be calculated, which is shown in Eq. 4:

$$w(U,S) = \frac{A_{|S|}^{|S|} A_{(n-1)-|S|}^{(n-1)-|S|}}{A_n^n} = \frac{|S|!(n-1-|S|)!}{n!} \tag{4}$$

where $w(U,S)$ is the weight for subset $S$; $n$ is the total number of features; $n!$ is the number of permutations of the complete set $U$; $|S|$ represents the number of features in the subset $S$; $|S|!$ is the number of permutations of the subset $S$; and ($n$-1-$|S|$) is the number of remaining features in the complete set $U$ after removing the features to be interpreted and the subset $S$, and the corresponding permutation number is ($n$-1-$|S|$)!.

Based on the value function in Eq. 2 and the weight in Eq. 4, the Shapley value of the feature to be interpreted under different scenarios (i.e., different feature combinations) can be evaluated, as shown in Eq. 5:

$$\begin{aligned}\phi_k &= \sum_{S \subseteq U \setminus \{X_k\}} w(U,S)\left[val(S \cup \{X_k\}) - val(S)\right] \\ &= \sum_{S \subseteq U \setminus \{X_k\}} \frac{|S|!(n-1-|S|)!}{n!}\left[val(S \cup \{X_k\}) - val(S)\right]\end{aligned} \tag{5}$$

where the feature to be interpreted is expressed as $X_k$, and its Shapley value is $\phi_k$; and $\left[val(S \cup \{X_k\}) - val(S)\right]$ denotes the value increment before and after the feature $X_k$ is added to the subset $S$.

The Shapley value has four excellent properties: efficiency, symmetry, dummy, and additivity, which are advantages that are not possessed by other feature impact analysis methods (Molnar, 2020; Roth, 1988; Shapley, 1951). The details of the four properties are presented in Appendix A.

**2.3 Ensemble randomized maximum likelihood and operational parameters optimization**

Based on the emulator and the Shapley value with different features at the single well level, InterOpt adopts the ensemble randomized maximum likelihood (EnRML) to optimize the operational parameters. The EnRML is a gradient-free data assimilation method, proposed in the field of history matching in petroleum engineering (Gu and Oliver, 2007). The essence of EnRML is to maximize the posterior probability through the Gauss-Newton method, and the gradients are replaced by covariance during optimization (Chang et al., 2017; Chen et al., 2019; Oliver et al., 2008). The architecture of EnRML is shown in Fig. 3.

The EnRML possesses advantages in small sample training, and is suitable for petroleum engineering where data acquisition is expensive and time-consuming. EnRML can also be combined with neural networks to solve problems in practice, such as ENN for tabular data in water-oil two-phase flow problem (Chen et al., 2019), EnLSTM for series data in well log generation problem (Chen and Zhang, 2020), and power load forecasting problem (Chen and Zhang, 2021). The EnRML is constructed based on Bayes' theorem. Its essence is to maximize the posterior probability, which is defined as follows:

$$p(m|d_{obs}) = \frac{p(m)p(d_{obs}|m)}{p(d_{obs})} \propto p(m)p(d_{obs}|m) \tag{6}$$

where $m$ denotes the model parameters; $d_{obs}$ denotes the observation; $p(m|d_{obs})$ is the posterior probability; $p(m)$ denotes the prior probability; and $p(d_{obs}|m)$ is the likelihood function.

Under the assumption that the observation is equivalent to the sum of the model predictions and stochastic errors and the errors obey a normal distribution, the objective function can be obtained by maximizing the posterior probability in Eq. 6, as shown in Eq. 7. The details about the objective function are provided in Supporting Information. The first term in Eq. 7 is called model mismatch, and it is proportional to the square of the difference between the model parameter and its prior estimate. The second term is defined as the data mismatch, and it is calculated based on the difference between the prediction and the observation.

$$m^* = \arg\min \left(\frac{1}{2}(g(m)-d_{obs})^T C_D^{-1}(g(m)-d_{obs}) + \frac{1}{2}(m-m_{pr})^T C_M^{-1}(m-m_{pr})\right) \tag{7}$$

Finally, the core of EnRML is the gradient-free update formula in the feedback process (Eq. 8), in which model mismatch and data mismatch are used to update between adjacent iterations. The update formula is detailed introduced in Supporting Information.

$$m_j^{l+1} = m_j^l - \frac{1}{1+\lambda_l}\left[C_{M_l} - C_{M_l,D_l}\left((1+\lambda_l)C_D + C_{D_l}\right)^{-1} C_{M_l,D_l}^T\right] C_M^{-1}(m_j^l - m_{pr,j})$$
$$- C_{M_l,D_l}\left((1+\lambda_l)C_D + C_{D_l}\right)^{-1}\left(g(m_j^l) - d_{obs,j}\right) \quad j=1,...,N_e \tag{8}$$

where $l$ and $j$ are the iteration and realization index, respectively; $g(m)$ is the estimation; $C$ denotes the cross-covariance matrix; $pr$ is the prior estimate; and the subscript M and D represent the model parameters and the estimations, respectively.

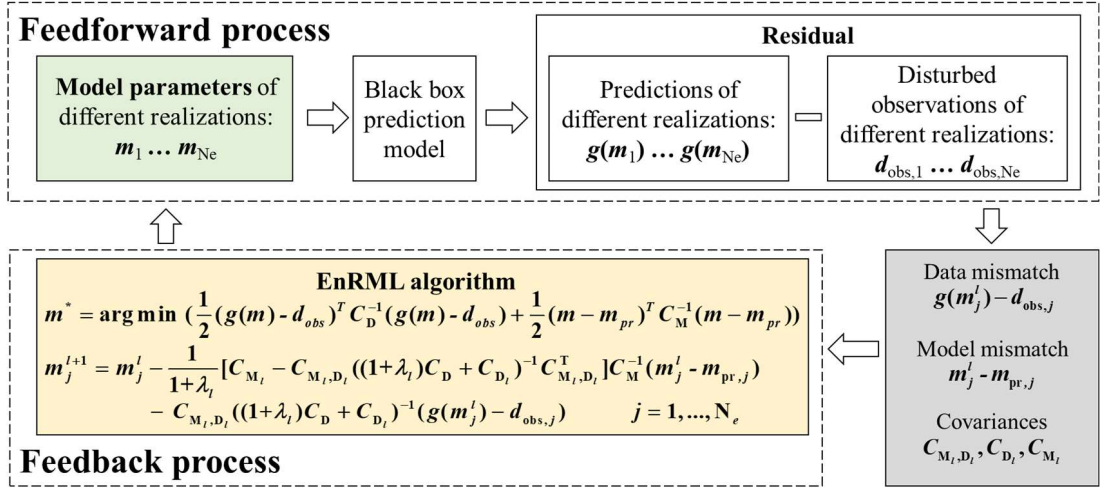

Fig. 3. Architecture of ensemble randomized maximum likelihood (EnRML).

In InterOpt, the input features include operational parameters and geological parameters. The EnRML can optimize the operational parameters on the premise of unchanged geological parameters. Specifically, the operational parameters are taken as the optimization objects (i.e., model parameters) of the EnRML, and the geological parameters are considered as a part of the black box prediction model in Fig. 3 since they are fixed in the drilling and fracturing process. It should be mentioned that the unchanged geological parameters do not mean that the geological parameters of all wells are the same, but rather that the geological parameters of each well remain unchanged in the optimization.

Due to the high cost and time-consuming process of obtaining shale gas development data, operational parameters optimization is a typical small-data problem, where the EnRML offers advantages over traditional methods since it can extract information from limited data more efficiently.

**2.4 Interpretable machine learning optimization (InterOpt)**

InterOpt consists of three parts, and the model architecture is shown in Fig. 4. The blue part is the neural network-based emulator describing the mapping relationship of real-world features. The green part analyzes the neural network through interpretable machine learning to evaluate the impact of each feature on the targets (e.g., average cost). The yellow part uses the EnRML to iteratively optimize the adjustable operational parameters to reduce the cost and increase the efficiency of shale gas development. Specifically, in order to optimize operational parameters via EnRML, the operational parameters, geological parameters, and outputs (e.g., average cost) are taken as the adjustable model parameters ($m$), fixed parameters, and target features in EnRML, respectively. The optimization objective of EnRML is to minimize the average cost or maximize production while the optimized drilling and fracturing plan are as close to the prior plan as possible.

It should be mentioned that the neural network prediction may contain negative values since the data are normalized. Therefore, the data mismatch term in EnRML (Eq. B.1 in Appendix B) cannot be directly used to minimize the average cost; otherwise, the output value will approach negative infinity, resulting in optimization failure. InterOpt solves this challenge by transforming the network output value into a bounded closed interval (please refer to Appendix B for details).

For different well conditions, the importance of each operational parameter varies. However, conventional optimization methods treat all features equally, and cannot provide customized optimization plans for each well. InterOpt adjusts the iterative optimization process based on the specific contributions of different features (i.e., the Shapley value of operational parameters) using dynamic weights to overcome this challenge, and its approach is detailed in Appendix C.

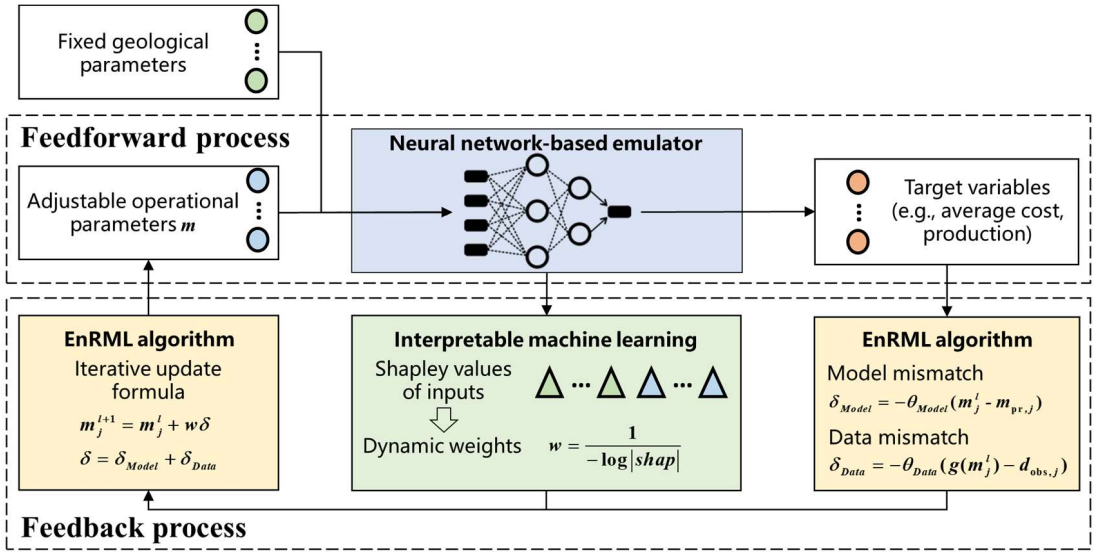

**Fig. 4.** Illustration of the iterative updating process of InterOpt.

In addition, InterOpt proposes two strategies to increase optimization efficiency and avoid training failure in practice: adaptive step size and block optimization, which are introduced in Appendix D. The primary goal of adaptive step size is to adjust the optimization process based on the model performance of each iteration step (i.e., self-adjusting). Furthermore, the block optimization assists InterOpt to avoid divergence in the optimization process by balancing exploration and exploitation.

**3. Experiments**
**3.1 Data description and experiment settings**

To test the performance of InterOpt, several experiments are carried out in this section. The operational parameters of 104 shale gas wells in the Sichuan Basin are optimized as a study case. The dataset contains a total of 13 features (12 operational and geological parameters as input features and one target feature), of which eight features can be changed in the drilling and fracturing process (i.e., operational parameters) and the other four features are fixed in the shale gas development. The 12 input features are shown in Table 2.

The target feature is the average cost (USD/m³), which is defined as the total cost of a single well divided by the estimated ultimate recovery (EUR). The total cost comes from the accounting report of each well, including pre-drilling cost (civil engineering), drilling cost (drilling bit cost, casing cost, top drive drilling cost, mud fluid service cost, geosteering service cost, environmental protection cost), cementing cost, logging cost, etc. It is worth mentioning that all 104 wells have been settled, which means that the total cost of each well can be accurately calculated.

**Table 2.** Input features of 104 shale gas wells in the Sichuan Basin study case.

| Adjustable operational parameters | | Fixed geological parameters |
|---|---|---|
| Content of acid fluid in drilling fluid (m³) | Content of ceramsite in fracturing fluid (t) | High-quality reservoir thickness (m) |
| Content of guanidine gum in fracturing fluid (m³) | Fracturing stages | Pressure coefficient |
| Content of slick water in fracturing fluid (m³) | Horizontal section length (m) | Depth (m) |
| Content of quartz sand in fracturing fluid (t) | Number of wells on the platform | Daily gas production of the first year (m³/day) |

In terms of neural network settings, InterOpt uses a three-layer fully connected neural network with 12 input features to form the emulator. The first hidden layer contains 20 neurons, the second hidden layer contains 10 neurons, and the output layer has one neuron. It is worth mentioning that the number of features and the number of wells available for training need to be considered when defining the network architecture. Indeed, it is not the case that a more complicated network is better. In real-world engineering applications, there is frequently a problem of insufficient data, and a too complicated network will result in overfitting. However, with the increase of problem complexity and training data, a reasonable increase in network complexity can assist to enhance model accuracy. For instance, a more complicated five-layer fully connected neural network with 50 neurons per layer was utilized in another study case which contains 43 features and 140 wells in Appendix E. In addition, the maximum training epoch is 500, and the Adam optimizer is used. In the feature impact analysis, InterOpt uses an interpretable machine learning algorithm called SHAP (Lundberg et al., 2020; Molnar, 2020) to calculate the Shapley value. This algorithm accelerates the calculation process of the Shapley value, which improves model efficiency.

In order to obtain a reliable experiment result, the leave-one-out cross validation method is used to evaluate the performance of InterOpt (Cawley, 2006; Evgeniou et al., 2004; Kohavi, 1995). Essentially, cross-validation is a resampling method that is especially useful when there are insufficient data. The process of cross-validation is as follows: first, the data are randomly divided into $k$ parts; second, one part is taken as the test set, and the remaining $k$-1 parts are used as the training set to obtain the model; and third, the model performance is evaluated on the test set. The above process is repeated $k$ times until each part of data has been evaluated once as a test set. Finally, the average of the $k$ evaluation results is taken as the performance of the model. When $k$ is the same as the total number of samples in the dataset, it is called leave-one-out cross-validation, which makes full use of all of the data and ensures the objectivity of the performance evaluation.

### 3.2 Emulator performance evaluation

The performance of the emulator is evaluated based on 104 shale gas wells in this section. Fig. 5 presents the scatterplot of the predictions of the neural network-based emulator versus their corresponding observations. Overall, the closer is the distribution of points to the 45° diagonal (blue line), the more precise are the predictions. Specifically, Fig. 5a shows the experiment results of the leave-one-out cross-validation, reflecting the prediction accuracy of the model. The fitness of the emulator to the training data is presented in Fig. 5b. The coefficient of determination ($R^2$) provides a measure of how well observed outcomes are replicated by the model, based on the proportion of total variation of outcomes explained by the model. The $R^2$ of the cross-validation results (Fig. 5a) and fitting results (Fig. 5b) are 0.71 and 0.85, respectively. When analyzing the feature impacts of emulators based on interpretable machine learning algorithms, the fitting model (i.e., Fig. 5b) can be used if the analysis is performed on wells that have already been produced. As shown in Fig. 5b, the emulator based on 104 wells has good fitness to the observations, which lays the foundation for subsequent optimization in InterOpt. It should also be mentioned that as the amount of training data increases, the accuracy of the emulator can be further improved.

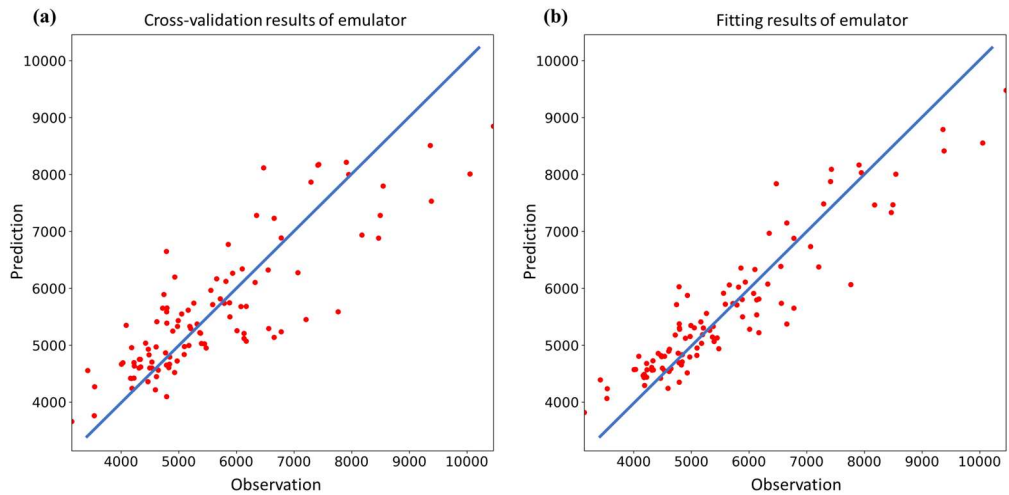

**Fig. 5.** Scatterplot of predictions and observations of the neural network-based emulator. (a) cross-validation results; (b) fitting results.

**3.3 Feature impact analysis**

Based on the neural network-based emulator, InterOpt calculated Shapley values for different features using the interpretable machine learning algorithm. The global Shapley value reflects the impact of each feature on the target feature in all wells, while the local Shapley value can measure the importance of each feature in different wells under different geological conditions.

**3.3.1 Local analysis for each well**

The results of feature impact analysis for Well 10, Well 50, and Well 100 in InterOpt are shown in Fig. 6. The red bars represent positive Shapley values, which means that the values of the corresponding features cause the average cost of the well to be higher than the mathematical expectation of all wells. The blue bars represent negative Shapley values, which indicates that the feature values result in lower average cost. In addition, the longer is the bar in the figure, the greater is the absolute value of the Shapley value, and the greater is the importance of the corresponding feature. A positive or negative Shapley value of a feature simply means that the current feature value causes the average cost of the well to be respectively larger or smaller than the mathematical expectation of all wells, and it does not reflect the positive or negative correlation between the feature and the average cost. For example, a small value of a feature that is positively correlated with average cost will also result in a negative Shapley value. Each bar in Fig. 6 shows the impact of the corresponding feature when taking the value in the current well. Therefore, the sum of the Shapley values for all features in one well represents the deviation between the average cost of that well and the mathematical expectation of the average cost of all wells. Specifically, if the sum of the Shapley values is negative, this means that the average cost of that well is lower than the mathematical expectation of all wells, and vice versa. The adjustable operational parameters are shown in bold font in Fig. 6.

Fig. 6 shows that the most important feature in Well 10 is the fracturing stages, and the red color of the bar indicates that the average cost tends to be higher than the mathematical expectation for the entire block since the well has 28 fracturing stages. Finally, the average cost of Well 10 is 0.57 CNY/m$^3$ (i.e., 0.090 USD/m$^3$). The main controlling factors of Well 50 are different from those of Well 10, where the most important feature is the content of slick water in fracturing fluid. This result indicates that Well 50 has a lower average cost because slick water consumption is only 32,067 t. In other words, under current geological conditions, reducing the amount of slick water has a smaller impact on production than on total costs. The top three main controlling factors of Well 50 also include horizontal section length and fracturing stages. Regarding Well 100, the average cost is approximately 0.45 CNY/m$^3$ (i.e., 0.071 USD/m$^3$), mainly because the well produces 144,800 m$^3$ of gas in the first year, and the high production reduces the average cost. Other main controlling factors for Well 100 include fracturing stages and the content of slick water in fracturing fluid. It can be seen from Fig. 6 that the main controlling factors of different wells are not the same, which requires different drilling and fracturing plans, and optimization strategies for different wells.

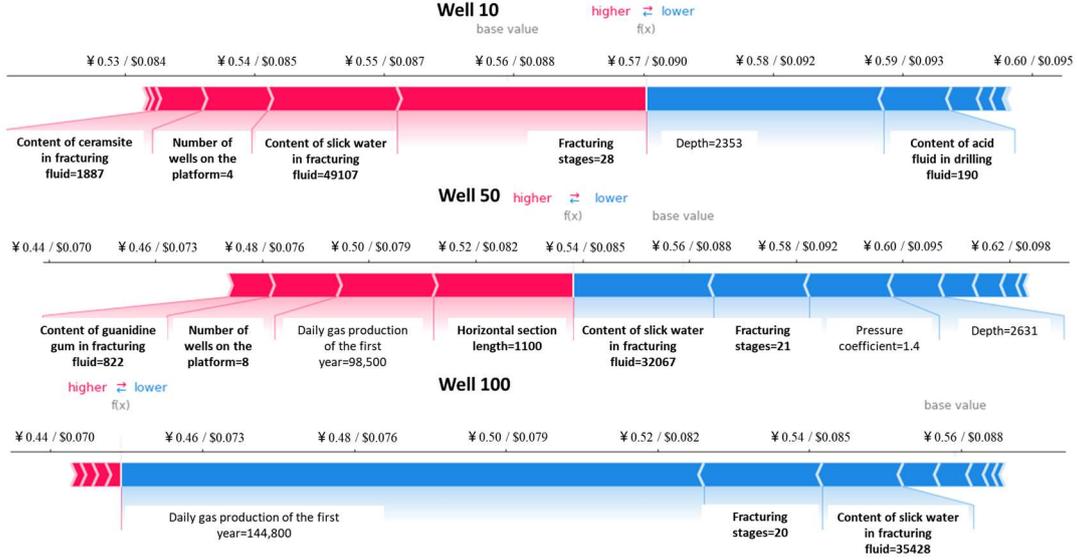

**Fig. 6.** Local Shapley value of Well 10, Well 50, and Well 100 in feature impact analysis.

### 3.3.2 Global analysis of the entire block

InterOpt can not only analyze the main controlling factors for a single well, but it can also provide global main controlling factors for the entire block, as shown in Fig. 7. The left side of Fig. 7 presents the bar chart of the global Shapley value (defined as Eq. 7), which is equal to the average of the absolute value of the local Shapley values in all wells, reflecting the global comprehensive impact of each feature on emulator output. Overall, the larger is the global Shapley value, the larger is the average expected marginal contribution of the feature, which indicates that the average cost is more sensitive to that feature in the block. Although the Shapley value is not equal to the sensitivity, it can effectively reflect the sensitivity of the feature. As can be seen from Fig. 7, for the average cost, the top-five global main controlling factors are: daily gas production of the first year; fracturing stages; the content of slick water in fracturing fluid; depth; and the content of guanidine gum in fracturing fluid.

$$shap_{global} = \sum_{i=1}^{N} |shap_i| / N \qquad (7)$$

where $shap_{global}$ denotes the global Shapley value; $shap_i$ denotes the local Shapley value of the $i^{th}$ well; and $N$ represents the number of wells.

The right side of Fig. 7 shows the scatterplot of the global Shapley value, which reflects the specific impact of different features on different wells. Each row in the figure corresponds to a feature, and each point corresponds to a well. Since the value of the feature is different in each well, the feature impact is also different (i.e., different positions in Fig. 7b). The color of the dot indicates the feature value in the current well. The position of the point on the abscissa corresponds to the Shapley value of the feature in the current well. Therefore, if the distribution of scatter points of a feature deviates greatly from the 0 value (i.e., the central axis), the influence of that feature is large. Moreover, the color of the scatter points reflects the correlation information. If the color of the points gradually changes from red to blue from left to right, this means that the feature is negatively related to the average cost (e.g., daily gas production of the first year). If the scatter points corresponding to the feature change from blue to red, it is a positive correlation (e.g., fracturing stages). If there is no regularity in the color change, the correlation is weak (e.g., the number of wells on the platform).

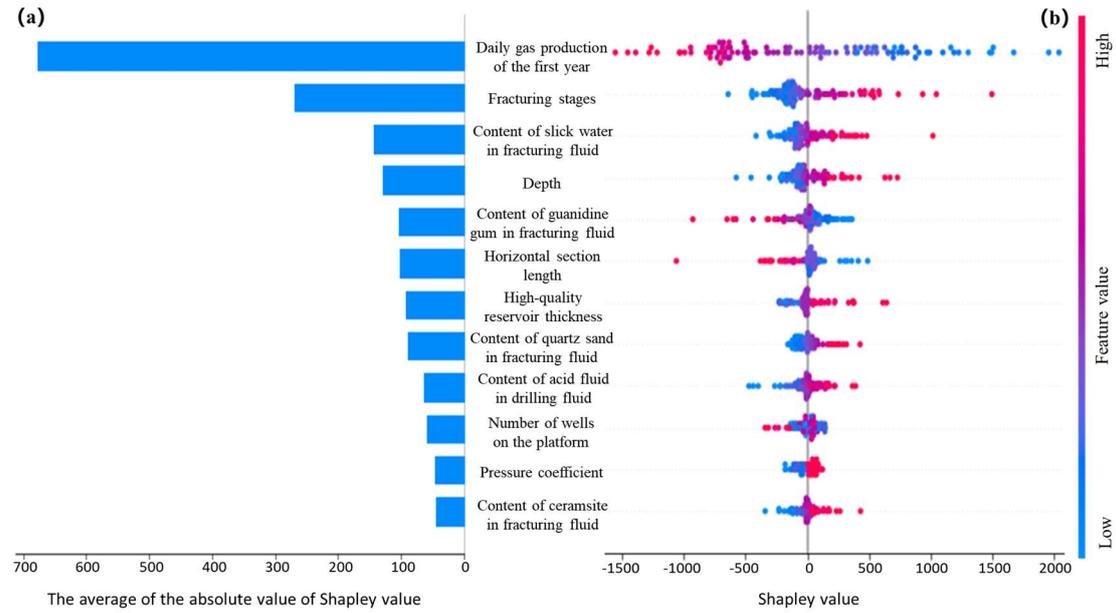

**Fig. 7.** Feature impact analysis results. (a) bar chart of global Shapley value; (b) scatterplot of global Shapley value.

In order to show the meaning and difference of Fig. 6 and Fig. 7, we take the content of acid fluid in drilling fluid as an example to demonstrate the method of analyzing Shapley values in InterOpt. This feature is ranked ninth in Fig. 7a, indicating that this feature is not one of the most important features at the block level (i.e., not a global main controlling factor). However, as shown in Fig. 6, it is one of the main controlling factors in Well 10. This difference shows that the main controlling factors of each well are related to their geological conditions, and thus the main controlling factors of a single well are not necessarily consistent with the global main controlling factors.

Therefore, each well possesses its own shortcomings, and the major contribution of InterOpt is to propose an optimization strategy for each well. According to Fig. 7b, it can be seen that the content of acid fluid in drilling fluid is positively correlated with the average cost. As a consequence, InterOpt tends to reduce the content of acid fluid in drilling fluid. However, since the global Shapley value of this feature is small (Fig. 7a), the amount of change of this feature in the optimization strategy is generally not large in practice, which is consistent with the experimental results of three example wells in Fig. 9, Fig. 10, and Fig. 11.

On the one hand, the global Shapley values comprehensively reflect the importance of features and provide correlation analysis results. On the other hand, the local Shapley values can reflect the influence of features in a single well, which establishes the foundation for operational parameters optimization at the single well level.

### 3.4 Operational parameters optimization results

In this section, eight adjustable operational parameters of 104 shale gas wells are optimized by InterOpt, and then the performance is evaluated by the neural network-based emulator. The results show that the average reduction rate of the 104 wells' average cost is 9.7%. The red bar chart in Fig. 8 shows the distribution of the average cost reduction. It can be seen that 33.7% of the wells (i.e., 35 wells) have an average cost reduction greater than 10%. In addition, the average cost reduction of 9.6% of the wells (i.e., 10 wells) is greater than 30%. The blue line in Fig. 8 represents the mean of the average cost reductions for the wells in each group. Experiments show that the existing operational parameters of approximately one-third of the wells (i.e., 31 wells) are close to the optimal drilling and fracturing plans (reduction rate less than 1%), which indicates that there is little room for further improvement. Approximately one-fourth of the wells have limited space for optimization, and the reduction rate is only 1% to 5%. Nevertheless, nearly half of the wells have large room for improvement (i.e., 46 wells with a reduction rate higher than 5%). Therefore, InterOpt can optimize the operational parameters of these wells under the premise of unchanged geological parameters, and provide different optimal drilling and fracturing plans for each well to reduce the average cost.

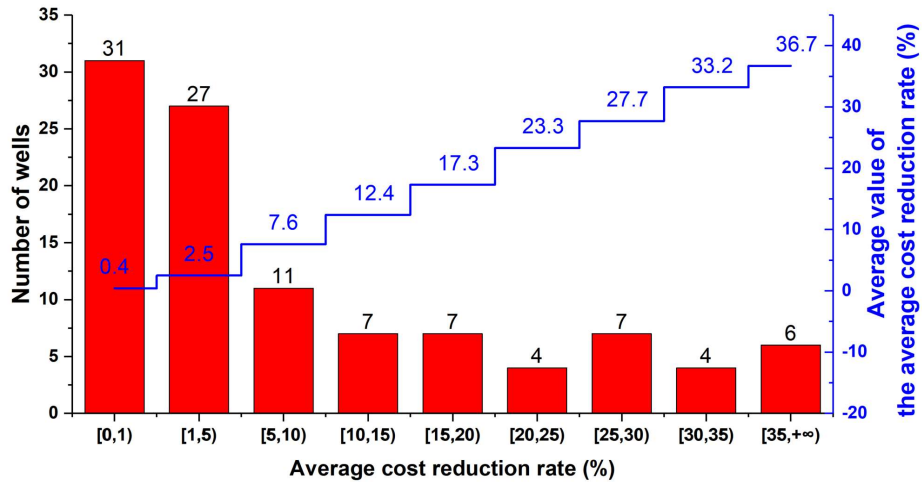

**Fig. 8.** Distribution plot of average cost reduction for InterOpt optimization results.

This study evaluates the effects of adaptive step size and block optimization (Appendix D) in InterOpt by comparing the results of ablation experiments. It can be seen from Table 3 that block optimization is crucial to ensure convergence of the model, and one-quarter to one-third of the wells (i.e., 25 to 33 wells, respectively) in the model without the block optimization strategy will diverge. Furthermore, the adaptive step size can automatically adapt the optimization process, which is conducive to balancing exploration and exploitation, and can improve the model performance.

**Table 3.** Effect of block optimization and adaptive step size in InterOpt.

| Training methods | | Number of wells that failed to optimize | | | Average cost reduction (%) |
|---|---|---|---|---|---|
| Block optimization | Adaptive step size | Not converged | No improvement | Total | |
| √ | √ | 0 | 3 | **3** | **9.7** |
| √ | | 0 | 25 | 25 | 9.0 |
| | √ | 25 | 0 | 25 | 7.1 |
| | | 33 | 0 | 33 | 8.5 |

In addition, three shale gas wells are taken as examples to demonstrate the optimization process and results of InterOpt. Their operational parameters before and after optimization are compared in Fig. 9, Fig. 10, and Fig. 11. Specifically, Fig. 9a shows that the average cost reduction for Well 10 gradually increased to 9.2% and converged as the iteration progressed. Fig. 9b shows the optimization process of the eight operational parameters in Well 10. Fig. 9c compares the values of the operational parameters before and after the optimization. It can be seen that the operational parameters are fine-tuned on the basis of the original plan. For Well 10, InterOpt mainly optimizes the fracturing stages. Ultimately, InterOpt reduced the average cost of Well 10 from 0.57 CNY/m$^3$ (i.e., 0.090 USD/m$^3$) to 0.52 CNY/m$^3$ (i.e., 0.081 USD/m$^3$).

In Well 10, InterOpt mainly optimized the fracturing stages; whereas, in other wells, InterOpt may adjust other operational parameters according to the specific geological conditions of the well. For example, Fig. 10 and Fig. 11 show the optimization process of Well 41 and Well 52, respectively, where InterOpt also optimizes the number of wells on the platform and the horizontal section length, in addition to adjusting the fracturing stages. It is worth noting that although the number of wells on the platform is not the global main controlling factor in Fig. 7, this feature plays an important role in Well 41, which indicates the significance of local interpretation in practice.

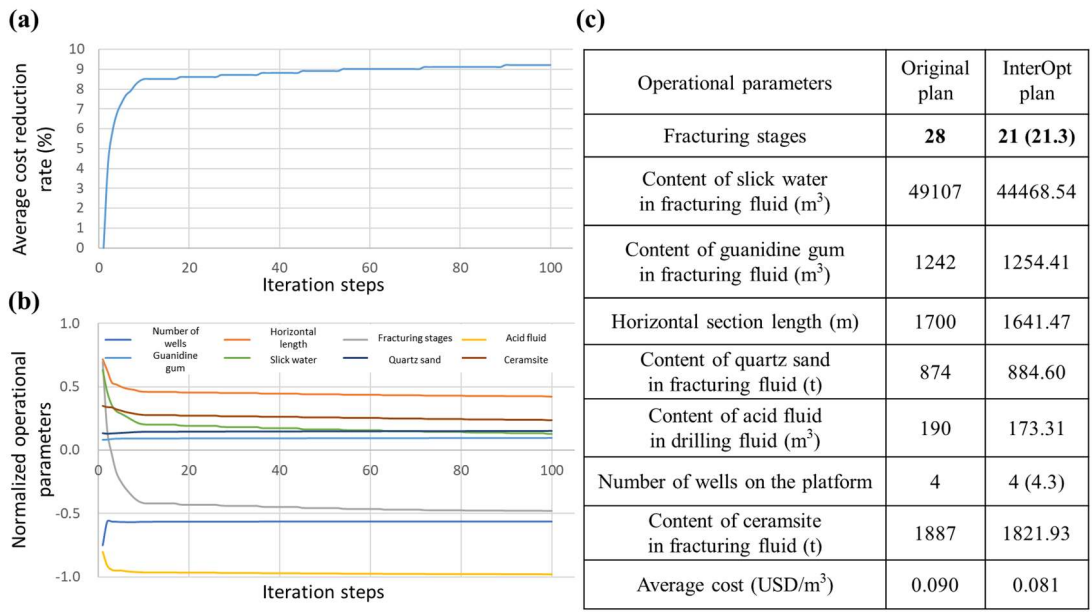

**Fig. 9.** Optimization process of InterOpt in Well 10. (a) average cost reduction curve; (b) optimization process of operational parameters; (c) operational parameters before and after optimization (the numbers in parentheses are the outputs of InterOpt, which are rounded up to the nearest decimal point).

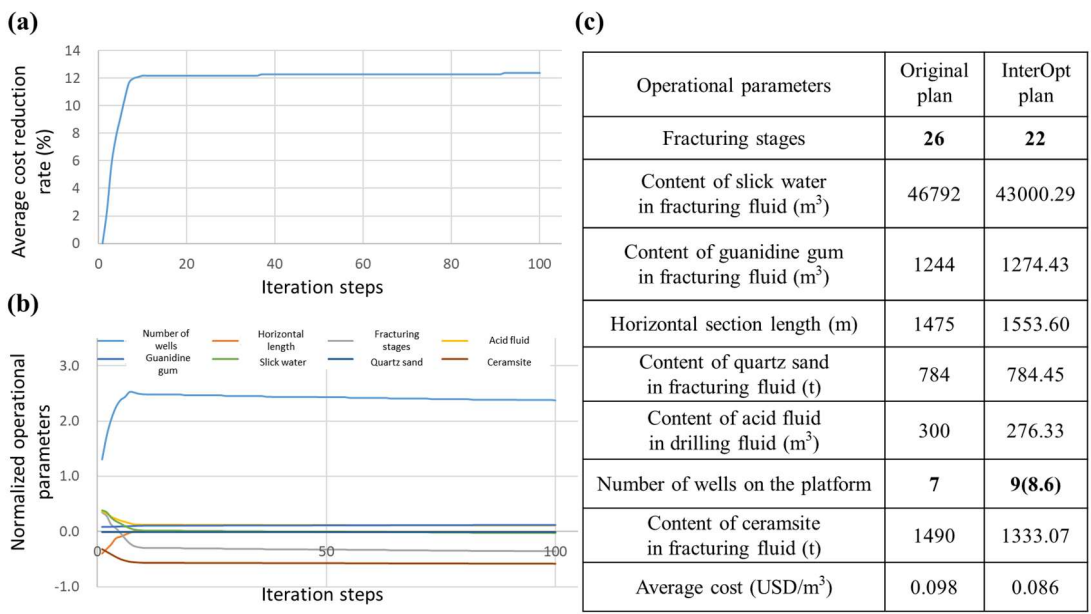

**Fig. 10.** Optimization process of InterOpt in Well 41. (a) average cost reduction curve; (b) optimization process of operational parameters; (c) operational parameters before and after optimization (the numbers in parentheses are the outputs of InterOpt, which are rounded up to the nearest decimal point).

In addition, Well 52 obtained two sets of plans during the optimization process, i.e., the in-process plan and the converged plan (Fig. 11c). The in-process plan is more aggressive, and the average cost reduction is higher, but it deviates further from the original plan and increases the engineering difficulty. For instance, although the average cost of the in-process plan is only 0.59 CNY/m$^3$ (i.e., 0.093 USD/m$^3$), it requires that the horizontal section length be increased to 2,432 m. However, this length is only 1,277 m in the converged plan, which is easier to implement. Moreover, the average cost of the converged plan is reduced to 0.63 CNY/m$^3$ (i.e., 0.099 USD/m$^3$), which is still lower than the original plan. In most practical situations, the easy-to-implement converged plan is chosen.

It should be mentioned that it is meaningless to simply reduce various operational parameters and ignore the impact on production. By optimizing average cost, we hope to find a balance between total cost and production, i.e., an "economic sweet spot" is sought. In fact, InterOpt does not directly reduce all operational parameters. For example, in Fig. 11c, InterOpt proposes to increase the horizontal section length by 23%, which indicates that although increasing the length will bring a higher total cost, the increase in production results in lower average cost.

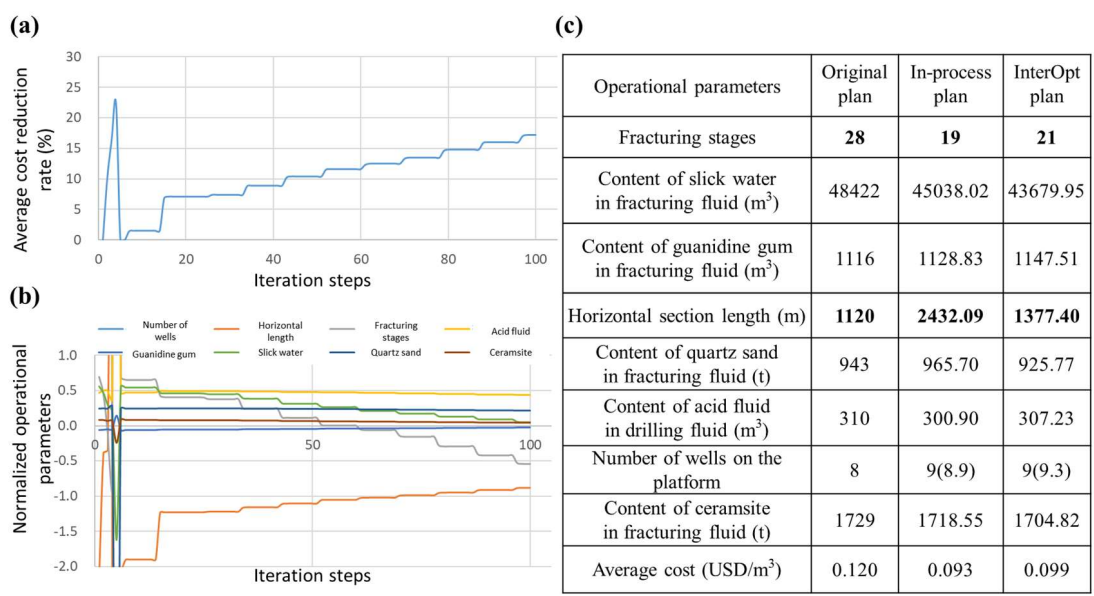

**Fig. 11.** Optimization process of InterOpt in Well 52. (a) average cost reduction curve; (b) optimization process of operational parameters; (c) operational parameters before and after optimization (the numbers in parentheses are the outputs of InterOpt, which are rounded up to the nearest decimal point).

The operational parameters optimized for different wells in Fig. 9 to Fig. 11 are not consistent, which also reflects the advantage of InterOpt over other methods, such as correlation or sensitivity analysis. InterOpt can provide different optimization plans for different wells based on their specific geological conditions via local interpretation.

It is worth noting that InterOpt's input features and optimization objectives can be flexibly altered to meet diverse needs and adapt to varied conditions in practice. Another study case based on 140 wells and 43 features is taken as an example to verify the transferability and scalability of InterOpt to different optimization objectives, the details of which are given in Appendix E.

**3.5 Inspiration from InterOpt optimization result**

InterOpt tends to lower the number of fracturing stages in the optimization process according to the results in section 3.4, which has attracted the interest of oilfield experts. This phenomenon reveals InterOpt's finding that the marginal contribution of the number of fracturing stages to production will decrease rapidly. Once the number of fracturing stages reaches a certain threshold, the increase in production cannot offset the rapid increase in cost and results in higher average cost (i.e., total cost of a single well divided by the EUR).

In order to verify this finding, oilfield experts analyzed and evaluated actual historical exploration and development data from the Changning Block in the Sichuan Basin. The data show that the number of fracturing stages of a single well generally has increased from 20 stages to more than 30 stages since 2017 (Fig. 12a). However, the estimated ultimate recovery (EUR) of a single well did not exhibit an upward trend, as shown in Fig. 12b. In other words, increasing the number of fracturing stages beyond 20, according to historical data, does not improve the well's performance. Therefore, the optimization strategy of InterOpt (i.e., reducing the number of fracturing stages) is consistent with the historical data, and is in accordance with experience in oil and gas development.

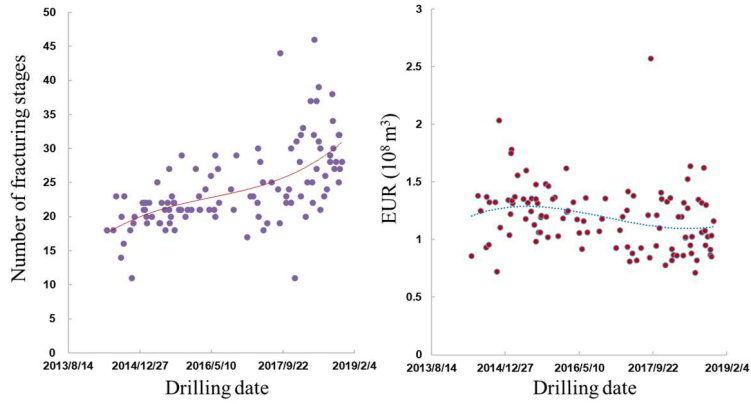

**Fig. 12.** (a) trend of the number of fracturing stages in horizontal wells; (b) trend of estimated ultimate recovery.

## 4. Conclusion

In this study, interpretable machine learning is applied for the first time to an operational parameters optimization problem in petroleum exploration and development. The method, named InterOpt, starts by modeling the mapping relationship between different physical parameters with neural networks (i.e., virtual environment construction), then uses interpretable machine learning to analyze the influence and importance of features at the single well level (i.e., feature impact analysis), and finally uses the ensemble randomized maximum likelihood algorithm to improve drilling and fracturing plans (i.e., operational parameters optimization).

InterOpt can optimize the shortcomings of each well according to specific geological conditions. In the experiment, InterOpt proposed different optimization plans for 104 wells, respectively, and achieved an average cost reduction of 9.7% by adjusting eight operational parameters, such as fracturing stages, content of slick water in fracturing fluid, and horizontal section length. In addition, the transferability and scalability of InterOpt to different optimization objectives is verified based on the optimization of test production in Appendix E, in which the final production increase rate is 16.9%.

The contributions of InterOpt include the following:
- In comparison to conventional main controlling factor methods, InterOpt can not only provide global main controlling factors, but it can also give local (i.e., single well level) main controlling factors to support different optimization of each well.
- InterOpt can evaluate the genuine contribution of each feature more comprehensively and fairly by taking into account the interaction and coupling between multiple features, as opposed to typical algorithms that assess the impact of each feature independently.
- InterOpt can optimize all adjustable operational parameters simultaneously, unlike existing algorithms that can only optimize a few main controlling factors. As a result, InterOpt's optimization is more thorough and flexible.
- Two optimization techniques, adaptive step size and block optimization, are proposed in Appendix D, both of which are critical for ensuring convergence of the optimization process. Meanwhile, the connection between the optimization process and the interpretable machine learning algorithm can be established using two mathematical methods, the transformation of neural network outputs (Appendix B) and the dynamic weights based on Shapley value (Appendix C), to realize optimization of operational parameters.

## 5. Discussion

The task of reducing costs and increasing efficiency by optimizing operational parameters in shale gas development is extremely challenging. This paper provides a feasible solution to this challenge based on interpretable machine learning. Because InterOpt is essentially a mathematical tool, different optimization objectives may be assigned to specific problems in different scenarios, and a different optimization strategy for each well can be used as a reference for well site engineers. The study case used in the experiment section is only to verify the effectiveness of InterOpt. Input and output features can be adjusted according to the needs of the well site engineer in practical applications.

The optimization of each well by InterOpt does not require retraining the neural network and has high computational efficiency. InterOpt is also compatible with both CPU and GPU computing. Despite the fact that GPU computation is faster, due to the restricted GPU resources at the well site, InterOpt is computed by default using the CPU. InterOpt takes approximately 3 min to optimize one well for 100 iterations based on an Intel Xeon Gold 5120 (workstation-class CPU). It takes 5 to 8 min using an Intel Core i5-10210U (laptop-class CPU).

Because the Shapley value calculation in the feature impact analysis (i.e., step 2 in Fig. 1) and the EnRML algorithm in the operational parameters optimization (i.e., step 3 in Fig. 1) have strict mathematical proofs that can ensure correctness, the accuracy of InterOpt is primarily determined by the emulator in virtual environment construction (i.e., step 1 in Fig. 1), which is essentially a data-driven neural network. The performance of the emulator in InterOpt is dependent on the quality and quantity of training data. If critical features are absent from the training data, InterOpt cannot guarantee its accuracy. The users are capable to assess whether the current data are sufficient to support the modeling via the cross-validation procedure described in section 2.2. It should be mentioned that since the neural network has theoretically infinite expressive ability, the accuracy of the emulator can be further improved with the accumulation of data in practice. Meanwhile, we will focus on small data learning and transfer learning to weaken this constraint in future studies.

It is also necessary to further explore the methods of embedding physical constraints into InterOpt since the optimization results of InterOpt may not adhere to the physical mechanisms. Potential physical constraints include restricting the data distribution of operational parameters to avoid results that violate physical rules, and combining the mathematical programming to optimize parameters (e.g., fracturing stages) to prevent the optimization results from being decimals (Zhang et al., 2017a; Zhang et al., 2017b). These topics will be explored in depth in future studies.

**Acknowledgement**

The idea of using the Shapley value in this work is motivated by the discussions with Haibin Chang and Wenbo Hu in 2019. The authors would like to thank them for their contributions during the discussion. In 2020, the authors verified the algorithm with actual data based on a project of Research Institute of Petroleum Exploration and Development of CNPC. In 2021 and 2022, InterOpt has been blindly tested, and it is further improved based on hundreds of wells from different blocks.

This work is funded by the National Natural Science Foundation of China (Grant No. 62106116), the Shenzhen Key Laboratory of Natural Gas Hydrates (Grant No. ZDSYS20200421111201738), and the SUSTech - Qingdao New Energy Technology Research Institute.

**Appendix A: Four properties of Shapley value: efficiency, symmetry, dummy, and additivity**

The following four properties of Shapley value are the advantages that are not possessed by other feature impact analysis methods [14, 20, 21].

- Efficiency means that the sum of all feature contributions of a sample is equal to the difference between the prediction of the sample and the mathematical expectation of all samples, i.e., Eq. A.1 is satisfied.

$$\sum_{j=1}^{n} \phi_j = f(X = x) - Ex(f(X)) \tag{A.1}$$

- Symmetry means that if two features $x_j$ and $x_k$ contribute equally to all possible feature combinations, their overall contribution should also be the same (i.e., their Shapley value is the same). According to Eq. 5, for any $S \subseteq \{x_1, ..., x_n\} \setminus \{x_j, x_k\}$, if Eq. A.2 is satisfied, then we have $\phi_j = \phi_k$ (i.e., the definition of symmetry).

$$val(S \cup \{x_j\}) = val(S \cup \{x_k\}) \tag{A.2}$$

- Dummy means that a feature should have a Shapley value of 0 if it does not affect the output of the model in any possible feature combination. According to Eq. 5, for any $S \subseteq \{x_1, ..., x_n\} \setminus \{x_j\}$, if Eq. C.3 is satisfied, then we have $\phi_j = 0$. Therefore, Shapley value has the property of dummy.

$$val(S \cup \{x_j\}) = val(S) \tag{A.3}$$

- Additivity means that when there are multiple gains in a game (i.e., the model has multiple outputs, or the model results depend on multiple sub-models, such as random forests), the contributions of the same feature to different outputs can be added. In other words, when there are multiple games, the distribution of the gains of one game does not affect the other games, which is obvious from Eq. 5.

**Appendix B: Neural network output values transformation method**

The output values of the neural network are transformed in InterOpt to avoid approaching negative infinity in the optimization process. Specifically, a hyperbolic tangent transformation is firstly applied to the outputs to ensure that the results are in the (-1, 1) value range. Then, the target value is set to a matrix with all elements of -1, so that the outputs of the neural network-based emulator can approach the lower limit of the value range and minimize the average cost in the optimization process.

Specifically, the original data mismatch term in EnRML is shown in Eq. B.1, and the transformed data mismatch term in InterOpt is shown in Eq. B.2. In addition to the data mismatch term, the optimization process of the EnRML also minimizes the model mismatch term (Eq. B.3), so that the optimized results are similar to the prior adjustable parameters (i.e., the original operational plan).

$$\delta_{Data}^{EnRML} = \omega_{Data}(g(m_j^l) - d_{obs,j}) \tag{B.1}$$

$$\delta_{Data}^{InterOpt} = \omega_{Data}(\tanh(g(m_j^l)) - (-1)) \tag{B.2}$$

$$\delta_{Model} = \omega_{Model}(m_j^l - m_{pr,j}) \tag{B.3}$$

where $\delta_{Data}^{EnRML}$ and $\delta_{Data}^{InterOpt}$ represent the data mismatch term in the EnRML and the transformed data mismatch term in InterOpt, respectively; $m_j^l$ and $g(m_j^l)$ denote the parameters to be optimized and the predictions of the $j^{th}$ realization in the $l^{th}$ iteration, respectively; $\tanh()$ represents the hyperbolic tangent function; $d_{obs,j}$ represents the observed value of the $j^{th}$ realization; $\delta_{Model}$ represents the model mismatch term; $m_{pr,j}$ represents the prior value of the adjustable parameter in the $j^{th}$ realization (e.g., the original operational plan); and $\omega_{Data}$ and $\omega_{Model}$ represent the coefficients of the data mismatch term and the model mismatch term in the EnRML, respectively, which are related to the covariance between observations $C_D$, the covariance between parameters and observations $C_{M,D}$, and the covariance between parameters $C_M$. The specific calculation process can be seen in EnRML (Gu and Oliver, 2007).

**Appendix C: Calculation method of dynamic weights based on Shapley value**

In the optimization process of the EnRML, each iterative step will generate a correction to update the adjustable parameters according to the loss function. In order to take advantage of the importance of different features, InterOpt employs an interpretable machine learning model to calculate the Shapley values of all of the adjustable parameters of the well to be optimized. Overall, the greater is the value, the more important is the corresponding feature in the existing feature combination of the well to be optimized, and its optimization should be prioritized.

Specifically, the dynamic weight $w$ is defined as Eq. B.1 in InterOpt. First, the absolute value is used to quantify the Shapley value ($shap$) since the positive or negative of the Shapley value has no influence on its importance. Then, the base 10 logarithm of the absolute value of the Shapley value ($\log|shap|$) is used to measure the magnitude. Finally, considering that most of the absolute value of the Shapley value is less than 1 and results in a negative magnitude, the reciprocal of the opposite number is taken as the dynamic weight to ensure that the weight is positively correlated with the importance of the corresponding feature.

$$w = \frac{1}{-\log|shap|} \tag{C.1}$$

Intuitively, the dynamic weight adjusts the corrections in the optimization process according to the importance of the feature to be optimized. Specifically, it amplifies the corrections of important features and weakens the corrections of other features. Eq. B.2 shows the dynamic weighted update formula in InterOpt. Compared with conventional methods that optimize all adjustable parameters equally, InterOpt focuses on optimizing important features under specific well conditions, which results in higher optimization efficiency in practice.

$$\begin{aligned} m_j^{l+1} &= m_j^l - w\delta \\ &= m_j^l + \frac{1}{\log|shap|}(\delta_{Model} + 10\delta_{Data}^{InterOpt}) \end{aligned} \tag{C.2}$$

where $w$ represents the dynamic weight matrix based on Shapley value; $shap$ represents the matrix composed of the Shapley values of all adjustable parameters; $m_j^{l+1}$ and $m_j^l$ denote the adjustable features (i.e., operational parameters) of the $l$<sup>th</sup> step and the $l+1$<sup>th</sup> step in the $j$<sup>th</sup> realization, respectively; $\delta$ is the corrections, which are calculated from the model mismatch term and the data mismatch term in Eq. B.2 and Eq. B.3.

**Appendix D: Adaptive step size and block optimization**

The adaptive step size is applied to adjusting the optimization process automatically based on the model performance of each iteration step). Specifically, if prediction performance improves after iteration (indicating that it is in the correct optimization direction), increasing the step size and using a more radical optimization strategy are conducive to faster convergence. If prediction performance declines after iteration (indicating that the optimization route is incorrect), it is better to reject the optimization, lower the step size, and use a more conservative optimization method to search for the proper optimization path. The flow chart of adaptive step size is presented in Fig. D.1a.

The block optimization in Fig. D.1b assists InterOpt to avoid divergence in the optimization process by balancing exploration and exploitation. The entire optimization procedure is divided into various blocks by InterOpt. The model's performance is permitted to deteriorate inside of each block, but at the end of each block the performance is assessed. If the model's performance increases, InterOpt moves on to the next block; otherwise, it rejects the update of this block and enters the next block. On the one hand, because block optimization permits model performance to degrade inside of a block, it prioritizes exploration, which is favorable to discovering new optimization routes and avoiding a local minimum. InterOpt, on the other hand, concentrates on the exploitation of current optimization directions at the block level, which helps to fully leverage the potential of existing strategies because the model's performance does not diminish between each block.

**Appendix E: Experiment on transferability and scalability to different optimization objectives**

In this appendix, the test production is optimized based on data from 140 wells in another block to verify that InterOpt can be applied to optimizing different objectives. The dataset covers 43 features, as shown in Table D.1. The adjustable and fixed parameters are determined according to the requirements of well site engineers, and thus some operational features are also regarded as fixed parameters. In practice, whether the inputs features are fixed in the optimization process can be flexibly altered to meet diverse needs and adapt to varied conditions.

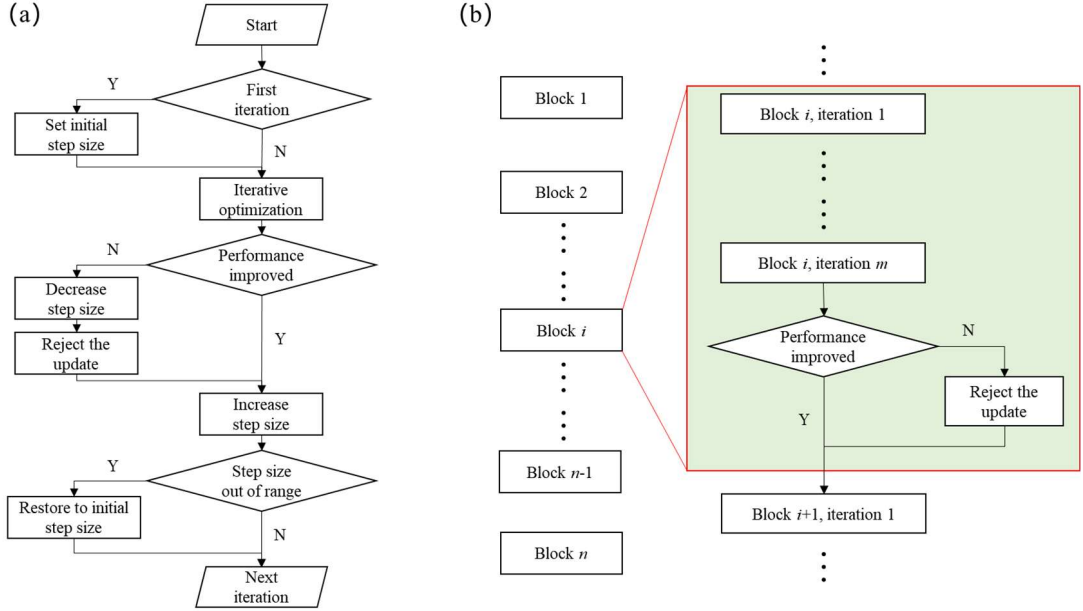

**Fig. D.1.** Training methods of InterOpt. (a) adaptive step size; (b) block optimization.

In this experiment, the operational parameters are optimized according to the geological conditions of each well with the goal of improving the test production. Fig. E.1 shows the 15 most important features (i.e., main controlling factors) affecting the test production, and Fig. E.2 shows the distribution of the test production improvement rate of all wells (which is similar to Fig. 8). The fact that 35% of the wells have an improvement rate of less than 5% indicates that the existing operational parameters are close to ideal plans for one-third of the wells. However, for the remaining wells, there is still room for improvement, and a total of 52 wells have tested production growth rates of more than 20%. Finally, the average test production increase rate of the 140 wells achieves 16.9%.

This experiment verifies the scalability of InterOpt, and demonstrates that it can work with a wide range of input parameters and accomplish different optimization tasks by changing the target. InterOpt is essentially a data-driven model, and its input and target features can be flexibly changed according to actual problem requirements.

**Table E.1.** Input features of 140 shale gas wells (20 adjustable features and 23 fixed features).

| Adjustable operational parameters | | Fixed parameters | |
|---|---|---|---|
| Fracturing stages | Average stage length (m) | Longitude | Latitude |
| Average sand volume per fracturing stage (t) | Average fluid volume per fracturing stage (m$^3$) | Casing pressure (MPa) | Average pump stop pressure (MPa) |
| Number of perforation clusters | Average sand volume per perforation cluster (t) | Formation fracture pressure (MPa) | Target point vertical depth difference (m) |
| Average fluid volume per perforation cluster (m$^3$) | Horizontal length (m) | Depth (m) | Well length in layer JC (m) |
| Total amount of fracturing fluid (m$^3$) | Total sand volume (t) | TOC | Gas saturation (m$^3$/t) |
| Well length in layer 1 (m) | Well length in layer 2 (m) | Porosity (%) | Permeability (mD) |
| Well length in layer 3 (m) | Well length in layer 4 (m) | Pressure coefficient | Sand-to-fluid ratio |
| Well length in layer 5 (m) | Well length in layer 6 to layer 9 (m) | Volume of acid fluid (m$^3$) | Volume of slick water (m$^3$) |
| 70-140 mesh proppant volume (t) | 40-70 mesh proppant volume (t) | Volume of guanidine gum (m$^3$) | Volume of pumped fluid (m$^3$) |
| 30-50 mesh proppant volume (t) | Angle between trajectory and minimum in-situ horizontal stress (°) | Total fluid volume in the well (m$^3$) | Fluid volume for handling abnormal conditions (m$^3$) |
| | | Number of leakages | Total leakage (m$^3$) |
| | | Nozzle size (mm) | |

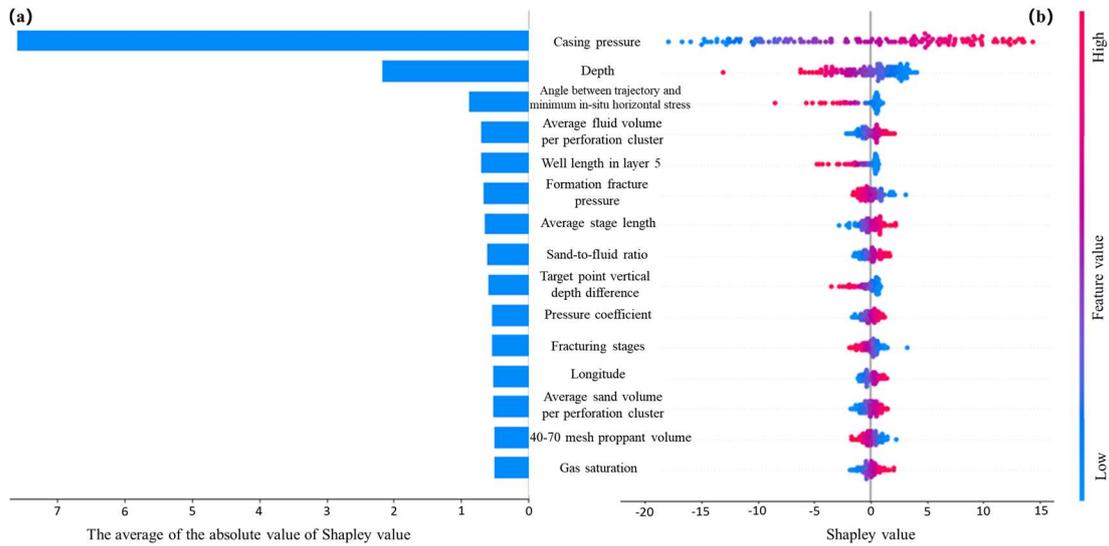

**Fig. E.1.** Feature impact analysis results. (a) bar chart of global Shapley value; (b) scatterplot of global Shapley value.

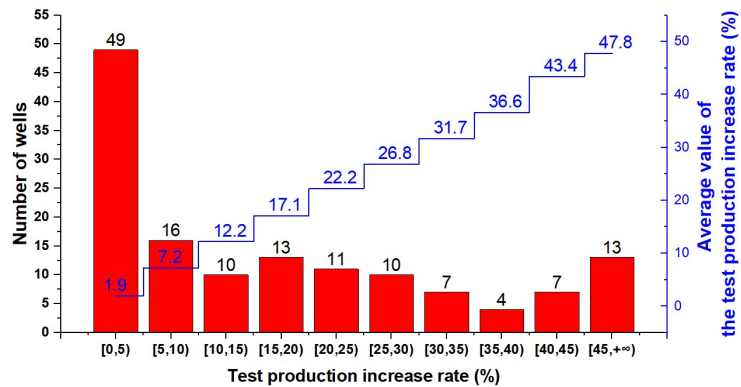

**Fig. E.2.** Distribution plot of test production increase rate for InterOpt optimization results.

**References**

Alvarez Melis, D. and Jaakkola, T., 2018. Towards robust interpretability with self-explaining neural networks. Advances in neural information processing systems, 31.

Aumann, R. and Hart, S., 2002. Handbook of game theory with economic applications, Elsevier.

Cawley, G.C., 2006. Leave-one-out cross-validation based model selection criteria for weighted LS-SVMs, The 2006 IEEE international joint conference on neural network proceedings. IEEE, pp. 1661-1668.

Chang, H., Liao, Q. and Zhang, D., 2017. Surrogate model based iterative ensemble smoother for subsurface flow data assimilation. Advances in Water Resources, 100: 96-108.

Chen, J., Song, L., Wainwright, M. and Jordan, M., 2018. Learning to explain: An information-theoretic perspective on model interpretation, International Conference on Machine Learning. PMLR, pp. 883-892.

Chen, Y., Chang, H., Meng, J. and Zhang, D., 2019. Ensemble Neural Networks (ENN): A gradient-free stochastic method. Neural Networks, 110: 170-185.

Chen, Y. and Zhang, D., 2020. Well log generation via ensemble long short-term memory (EnLSTM)

network. Geophysical Research Letters, 47(23): e2020GL087685.

Chen, Y. and Zhang, D., 2021. Theory-guided deep-learning for electrical load forecasting (TgDLF) via ensemble long short-term memory. Advances in Applied Energy, 1: 100004.

Cipolla, C.L., Lolon, E., Erdle, J. and Tathed, V.S., 2009. Modeling well performance in shale-gas reservoirs, SPE/EAGE Reservoir characterization and simulation conference. OnePetro.

Cybenko, G., 1989. Approximation by superpositions of a sigmoidal function. Mathematics of Control, Signals Systems, 2(4): 303-314.

Evgeniou, T., Pontil, M. and Elisseeff, A., 2004. Leave one out error, stability, and generalization of voting combinations of classifiers. Machine Learning, 55(1): 71-97.

Gu, Y. and Oliver, D.S., 2007. An iterative ensemble Kalman filter for multiphase fluid flow data assimilation. SPE Journal, 12(04): 438-446.

Guo, X., Hu, D., Li, Y., Wei, Z., Wei, X. and Liu, Z., 2017. Geological factors controlling shale gas enrichment and high production in Fuling shale gas field. Petroleum Exploration and Development, 44(4): 481-491.

Hornik, K., 1991. Approximation capabilities of multilayer feedforward networks. Neural Networks, 4(2): 251-257.

Huang, L.-T., Gromiha, M.M. and Ho, S.-Y., 2007. iPTREE-STAB: interpretable decision tree based method for predicting protein stability changes upon mutations. Bioinformatics, 23(10): 1292-1293.

Kim, B., Wattenberg, M., Gilmer, J., Cai, C., Wexler, J. and Viegas, F., 2018. Interpretability beyond feature attribution: Quantitative testing with concept activation vectors (tcav), International conference on machine learning. PMLR, pp. 2668-2677.

Kohavi, R., 1995. A study of cross-validation and bootstrap for accuracy estimation and model selection, International joint conference on artificial intelligence. Montreal, Canada, pp. 1137-1145.

Li, D., Fu, M., Huang, Y., Wu, D. and Xue, R., 2021. The characteristics and main controlling factors for the formation of micropores in shale from the Niutitang Formation, Wenshuicun Section, Southwest China. Energies, 14(23): 7858.

Liang, Y., Li, S., Yan, C., Li, M. and Jiang, C., 2021. Explaining the black-box model: A survey of local interpretation methods for deep neural networks. Neurocomputing, 419: 168-182.

Liu, W., Liu, W.D. and Gu, J., 2020. Forecasting oil production using ensemble empirical model decomposition based Long Short-Term Memory neural network. Journal of Petroleum Science Engineering, 189: 107013.

Lundberg, S.M., Erion, G., Chen, H., DeGrave, A., Prutkin, J.M., Nair, B., Katz, R., Himmelfarb, J., Bansal, N. and Lee, S.-I., 2020. From local explanations to global understanding with explainable AI for trees. Nature Machine Intelligence, 2(1): 56-67.

Lundberg, S.M. and Lee, S.-I., 2017. A unified approach to interpreting model predictions, Proceedings of the 31st international conference on neural information processing systems, pp. 4768-4777.

Luo, G., Tian, Y., Bychina, M. and Ehlig-Economides, C., 2019. Production-strategy insights using machine learning: application for Bakken Shale. SPE Reservoir Evaluation Engineering, 22(03): 800-816.

Luo, S.-H., Xiao, L.-Z., Jin, Y., Liao, G.-Z., Xu, B.-S., Zhou, J. and Liang, C., 2022. A machine learning framework for low-field NMR data processing. Petroleum Science.

Molnar, C., 2020. Interpretable machine learning. Lulu. com.

Oliver, D.S., Reynolds, A.C. and Liu, N., 2008. Inverse theory for petroleum reservoir characterization and history matching.

Ribeiro, M.T., Singh, S. and Guestrin, C., 2016. " Why should i trust you?" Explaining the predictions of any classifier, Proceedings of the 22nd ACM SIGKDD international conference on knowledge discovery and data mining, pp. 1135-1144.

Rogers, S.J., Fang, J., Karr, C. and Stanley, D., 1992. Determination of lithology from well logs using a neural network. AAPG bulletin, 76(5): 731-739.

Roth, A.E., 1988. The Shapley value: essays in honor of Lloyd S. Shapley. Cambridge University Press.

Shapley, L.S., 1951. Notes on the N-person Game. Rand Corporation.

Shen, F., Ren, S.S., Zhang, X.Y., Luo, H.W. and Feng, C.M., 2021. A digital twin-based approach for optimization and prediction of oil and gas production. Mathematical Problems in Engineering, 2021.

Shwartz-Ziv, R. and Tishby, N., 2017. Opening the black box of deep neural networks via information. arXiv preprint arXiv:.00810.

Song, X., Liu, Y., Xue, L., Wang, J., Zhang, J., Wang, J., Jiang, L. and Cheng, Z., 2020. Time-series well performance prediction based on Long Short-Term Memory (LSTM) neural network model. Journal of Petroleum Science Engineering, 186: 106682.

Sun, H., Chawathe, A., Hoteit, H., Shi, X. and Li, L., 2015. Understanding shale gas flow behavior using numerical simulation. SPE Journal, 20(01): 142-154.

Tang, J., Fan, B., Xiao, L., Tian, S., Zhang, F., Zhang, L. and Weitz, D., 2021. A new ensemble machine-learning framework for searching sweet spots in shale reservoirs. SPE Journal, 26(01): 482-497.

Wang, S. and Chen, S., 2019. Insights to fracture stimulation design in unconventional reservoirs based on machine learning modeling. Journal of Petroleum Science Engineering, 174: 682-695.

Wang, S., Chen, Z. and Chen, S., 2019. Applicability of deep neural networks on production forecasting in Bakken shale reservoirs. Journal of Petroleum Science Engineering, 179: 112-125.

Yao, J., Sun, H., Fan, D.-y., Wang, C.-c. and Sun, Z.-x., 2013. Numerical simulation of gas transport mechanisms in tight shale gas reservoirs. Petroleum Science, 10(4): 528-537.

Zhang, D., Chen, Y. and Meng, J., 2018. Synthetic well logs generation via Recurrent Neural Networks. Petroleum Exploration Development, 45(4): 629-639.

Zhang, H., Liang, Y., Zhang, W., Wang, B., Yan, X. and Liao, Q., 2017a. A unified MILP model for topological structure of production well gathering pipeline network. Journal of Petroleum Science Engineering, 152: 284-293.

Zhang, H., Liang, Y., Zhou, X., Yan, X., Qian, C. and Liao, Q., 2017b. Sensitivity analysis and optimal operation control for large-scale waterflooding pipeline network of oilfield. Journal of Petroleum Science Engineering, 154: 38-48.

Zhang, Q., Yang, Y., Ma, H. and Wu, Y.N., 2019. Interpreting cnns via decision trees, Proceedings of the IEEE/CVF conference on computer vision and pattern recognition, pp. 6261-6270.

Zhao, X., Zhang, K., Chen, G., Xue, X., Yao, C., Wang, J., Yang, Y., Zhao, H. and Yao, J., 2020. Surrogate-assisted differential evolution for production optimization with nonlinear state constraints. Journal of Petroleum Science Engineering, 194: 107441.

# Supporting Information

# Interpretable machine learning optimization (InterOpt) for operational parameters: a case study of highly-efficient shale gas development

Yuntian Chen [1], Dongxiao Zhang [2,3,*], Qun Zhao [4], and Dexun Liu [4]

1. *Eastern Institute for Advanced Study, Yongriver Institute of Technology, Zhejiang, P. R. China*
2. *Department of Mathematics and Theories, Peng Cheng Laboratory, Guangdong, P. R. China*
3. *National Center for Applied Mathematics Shenzhen (NCAMS), Southern University of Science and Technology, Guangdong, P. R. China*
4. *Research Institute of Petroleum Exploration and Development, CNPC, Beijing, P. R. China*

* Corresponding author

As mentioned in the article, the ensemble randomized maximum likelihood (EnRML) for inverse problems is used as the update algorithm. The EnRML is constructed based on Bayes' theorem. Its essence is to maximize the posterior probability. The posterior is defined as follows:

$$p(m\,|\,d_{obs}) = \frac{p(m)p(d_{obs}\,|\,m)}{p(d_{obs})} \propto p(m)p(d_{obs}\,|\,m) \tag{S1}$$

where $m$ is the model parameters; $d_{obs}$ denotes the observed data; $p(m\,|\,d_{obs})$ is the posterior probability; $p(m)$ denotes the prior probability; and $p(d_{obs}\,|\,m)$ is the likelihood function.

If the observation is equivalent to the sum of the model predictions and stochastic errors that obey a normal distribution (Eq. S2)), the likelihood function should be equivalent to the probability of the error (Eq. S3)):

$$d_{obs} = g(m) + \varepsilon \tag{S2}$$

$$p(d_{obs}\,|\,m) = p(d_{obs} - g(m)) = p(\varepsilon) \tag{S3}$$

where $g(m)$ is the prediction; and $\varepsilon$ denotes a normally-distributed random vector, with mean 0 and covariance matrix $C_D$.

Since the error obeys a normal distribution, $p(\varepsilon)$ can be calculated according to the multivariate normal distribution (Eq. S4), and Eq. S3 can be rewritten as Eq. S5:

$$p(\varepsilon) \propto \exp[-\frac{1}{2}(\varepsilon\text{-}0)^T C_D^{-1}(\varepsilon\text{-}0)] \tag{S4}$$

$$p(d_{obs}|m) \propto \exp[-\frac{1}{2}(d_{obs} - g(m))^T C_D^{-1}(d_{obs} - g(m))] \tag{S5}$$

Regarding the prior probability, since the model parameters are assumed to be Gaussian variables, it can be obtained by:

$$p(m) \propto \exp[-\frac{1}{2}(m - m_{pr})^T C_M^{-1}(m - m_{pr})] \tag{S6}$$

where $m_{pr}$ and $C_M$ denote the prior estimate and the prior covariance of the model parameters, respectively.

Finally, the posterior probability distribution is calculated according to Bayes' theorem Eq. S1, as shown in Eq. S7. The first term in Eq. S7 is called model mismatch, and it is proportional to the square of the difference between the model parameter and its prior estimate. The second term is defined as the data mismatch, and it is calculated based on the difference between the prediction and the observation:

$$\begin{aligned} p(m|d_{obs}) &\propto p(d_{obs}|m)p(m) \\ &\propto \exp[-\frac{1}{2}(g(m) - d_{obs})^T C_D^{-1}(g(m) - d_{obs}) - \frac{1}{2}(m - m_{pr})^T C_M^{-1}(m - m_{pr})] \\ &\propto \exp[-O(m)] \end{aligned} \tag{S7}$$

where $O(m)$ is defined as the objective function, and it is proportional to the posterior probability; and $C_M$ and $C_D$ denote covariance matrixes, and they are used to balance the scale of the two mismatches so that they are on the same scale.

In the EnRML, the posterior probability is maximized to update the model parameters, which is equivalent to minimizing the objective function $O(m)$. The iterative update formula can be obtained by the Gauss-Newton method (equation (S1.8)) (Bertsekas, 1999; Chen and Oliver, 2013):

$$\begin{aligned} m^{l+1} &= m^l - H(m^l)^{-1} \nabla O(m^l) \\ &= m^l - \left[(1+\lambda_l)C_M^{-1} + G_l^T C_D^{-1} G_l\right]^{-1} \left[C_M^{-1}(m^l - m_{pr}) + G_l^T C_D^{-1}(g(m^l) - d_{obs})\right] \end{aligned} \tag{S8}$$

where $l$ denotes the iteration index; $H(m^l)$ is the modified Gauss-Newton Hessian matrix with the form $\left[(1+\lambda_l)C_M^{-1} + G_l^T C_D^{-1} G_l\right]$; $G_l$ denotes the sensitivity matrix; and $\lambda$ is a multiplier to mitigate the influence of large data mismatch (Li et al., 2003).

The calculation of the inverse of the matrix $((1+\lambda_l)C_M^{-1} + G_l^T C_D^{-1} G_l)$ with size $N_m \times N_m$

is required in Eq. S8, where $N_m$ is the number of model parameters. Since we are solving a small-data problem, $N_m$ is always larger than the number of data points ($N_d$). Therefore, to reduce computation complexity, Eq, S8 is reformulated to Eq. S10 with an inverse of a $N_d \times N_d$ matrix based on two equivalent equations (Eq. S9) (Golub and Van Loan, 2012):

$$\begin{aligned}(C_M^{-1}+G_l^T C_D^{-1} G_l)^{-1} &= C_M - C_M G_l^T (C_D + G_l C_M G_l^T)^{-1} G_l C_M \\ (C_M^{-1}+G_l^T C_D^{-1} G_l)^{-1} G_l^T C_D^{-1} &= C_M G_l^T (C_D + G_l C_M G_l^T)^{-1}\end{aligned} \quad (S9)$$

$$\begin{aligned}m^{l+1} = m^l &- \frac{1}{1+\lambda_l}\left[C_M - C_M G_l^T \left((1+\lambda_l)C_D + G_l C_M G_l^T\right)^{-1} G_l C_M\right] C_M^{-1}(m^l - m_{pr}) \\ &- C_M G_l^T \left((1+\lambda_l)C_D + G_l C_M G_l^T\right)^{-1} \left(g(m^l) - d_{obs}\right)\end{aligned} \quad (S10)$$

Furthermore, we can generate a group of realizations of the model parameters (Oliver et al., 2008), and replace the sensitivity matrixes by covariance and cross-covariance based on the following approximations Eq. S11) (Reynolds et al., 2006; Zhang, 2001). In addition, the final update formula is shown as Eq. S12:

$$\begin{aligned}C_{M_l,D_l} &\approx C_{M_l} \overline{G}^T \\ C_{D_l} &\approx \overline{G} C_{M_l} \overline{G}^T\end{aligned} \quad (S11)$$

$$\begin{aligned}m_j^{l+1} = m_j^l &- \frac{1}{1+\lambda_l}\left[C_{M_l} - C_{M_l,D_l}\left((1+\lambda_l)C_D + C_{D_l}\right)^{-1} C_{M_l,D_l}^T\right] C_M^{-1}(m_j^l - m_{pr,j}) \\ &- C_{M_l,D_l}\left((1+\lambda_l)C_D + C_{D_l}\right)^{-1}\left(g(m_j^l) - d_{obs,j}\right) \quad j=1,...,N_e\end{aligned} \quad (S12)$$

where $j$ denotes the realization index; $C_{M_l}$ is the covariance matrix of the updated model parameters at the $l^{th}$ iteration step; $C_M$ denotes the prior model variable covariance, which does not change with iterations; $\overline{G}_l$ is the average sensitivity matrix; $d_{obs,j}$ denotes a perturbed observation sampled from a multivariate Gaussian distribution, with mean $d_{obs}$ and covariance $C_D$; $N_e$ represents the number of realizations; $C_{M_l,D_l}$ denotes the cross-covariance between the updated model parameters $m$ and the prediction $g(m)$ at iteration step $l$ based on the ensemble of realizations; and $C_{D_l}$ is the covariance of predictions.

---
**Algorithm S1.** Minimize the objective function in the EnRML (Chen et al., 2019).
---
**Input:** $x$ and $y$

**Trainable parameter:** $m$

**Hyper-parameters:** $m_{pr}$, $C_D$, and $C_M$ (determined based on prior information)

For $j = 1,...,N_e$

1. Generate realizations of measurement error $\varepsilon$ based on its probability distribution function (PDF);

2. Generate initial realizations of the model parameters $m_j$ based on prior PDF;

3. Calculate the observed data $d_{obs}$ by adding the measurement error $\varepsilon$ to the target value $y$;

**Repeat**

    **Step 1:** Compute the predicted data $g(m_j)$ for each realization based on the model parameters;

    **Step 2:** Update the model parameters $m_j$ according to equation (S1.12). The $C_{M_l,D_l}$ and $C_{D_l}$ are calculated among the ensemble of realizations. Therefore, the ensemble of realizations is updated simultaneously;

**until** the training loss has converged.
---

The EnRML can be summarized as Algorithm S1. It should be mentioned that the calculation of derivatives is not required in the EnRML, and most variables in Eq. S12 are easily accessible statistics.

**References**

Alvarez Melis, D. and Jaakkola, T., 2018. Towards robust interpretability with self-explaining neural networks. Advances in neural information processing systems, 31.
Aumann, R. and Hart, S., 2002. Handbook of game theory with economic applications, Elsevier.
Bertsekas, D.P., 1999. Nonlinear Programming. Athena Scientific, Belmont.
Cawley, G.C., 2006. Leave-one-out cross-validation based model selection criteria for weighted LS-SVMs, The 2006 IEEE international joint conference on neural network proceedings. IEEE, pp. 1661-1668.
Chang, H., Liao, Q. and Zhang, D., 2017. Surrogate model based iterative ensemble smoother for subsurface flow data assimilation. Advances in Water Resources, 100: 96-108.
Chen, J., Song, L., Wainwright, M. and Jordan, M., 2018. Learning to explain: An information-theoretic perspective on model interpretation, International Conference on Machine Learning. PMLR, pp. 883-892.
Chen, Y., Chang, H., Meng, J. and Zhang, D., 2019. Ensemble Neural Networks (ENN): A gradient-free stochastic method. Neural Networks, 110: 170-185.
Chen, Y. and Oliver, D.S., 2013. Levenberg–Marquardt forms of the iterative ensemble smoother for efficient history matching and uncertainty quantification. Computational Geosciences, 17(4): 689-703.
Chen, Y. and Zhang, D., 2020. Well log generation via ensemble long short-term memory (EnLSTM) network. Geophysical Research Letters, 47(23): e2020GL087685.
Chen, Y. and Zhang, D., 2021. Theory-guided deep-learning for electrical load forecasting (TgDLF) via ensemble long short-term memory. Advances in Applied Energy, 1: 100004.
Cipolla, C.L., Lolon, E., Erdle, J. and Tathed, V.S., 2009. Modeling well performance in shale-gas reservoirs, SPE/EAGE Reservoir characterization and simulation conference. OnePetro.

Cybenko, G., 1989. Approximation by superpositions of a sigmoidal function. Mathematics of Control, Signals Systems, 2(4): 303-314.
Evgeniou, T., Pontil, M. and Elisseeff, A., 2004. Leave one out error, stability, and generalization of voting combinations of classifiers. Machine Learning, 55(1): 71-97.
Golub, G. and Van Loan, C., 2012. Matrix Computations. MD: Johns Hopkins University Press, Baltimore.
Gu, Y. and Oliver, D.S., 2007. An iterative ensemble Kalman filter for multiphase fluid flow data assimilation. SPE Journal, 12(04): 438-446.
Guo, X., Hu, D., Li, Y., Wei, Z., Wei, X. and Liu, Z., 2017. Geological factors controlling shale gas enrichment and high production in Fuling shale gas field. Petroleum Exploration and Development, 44(4): 481-491.
Hornik, K., 1991. Approximation capabilities of multilayer feedforward networks. Neural Networks, 4(2): 251-257.
Huang, L.-T., Gromiha, M.M. and Ho, S.-Y., 2007. iPTREE-STAB: interpretable decision tree based method for predicting protein stability changes upon mutations. Bioinformatics, 23(10): 1292-1293.
Kim, B., Wattenberg, M., Gilmer, J., Cai, C., Wexler, J. and Viegas, F., 2018. Interpretability beyond feature attribution: Quantitative testing with concept activation vectors (tcav), International conference on machine learning. PMLR, pp. 2668-2677.
Kohavi, R., 1995. A study of cross-validation and bootstrap for accuracy estimation and model selection, International joint conference on artificial intelligence. Montreal, Canada, pp. 1137-1145.
Li, D., Fu, M., Huang, Y., Wu, D. and Xue, R., 2021. The characteristics and main controlling factors for the formation of micropores in shale from the Niutitang Formation, Wenshuicun Section, Southwest China. Energies, 14(23): 7858.
Li, R., Reynolds, A.C. and Oliver, D.S., 2003. History matching of three-phase flow production data. SPE Journal, 8(4): 328–340.
Liang, Y., Li, S., Yan, C., Li, M. and Jiang, C., 2021. Explaining the black-box model: A survey of local interpretation methods for deep neural networks. Neurocomputing, 419: 168-182.
Liu, W., Liu, W.D. and Gu, J., 2020. Forecasting oil production using ensemble empirical model decomposition based Long Short-Term Memory neural network. Journal of Petroleum Science Engineering, 189: 107013.
Lundberg, S.M., Erion, G., Chen, H., DeGrave, A., Prutkin, J.M., Nair, B., Katz, R., Himmelfarb, J., Bansal, N. and Lee, S.-I., 2020. From local explanations to global understanding with explainable AI for trees. Nature Machine Intelligence, 2(1): 56-67.
Lundberg, S.M. and Lee, S.-I., 2017. A unified approach to interpreting model predictions, Proceedings of the 31st international conference on neural information processing systems, pp. 4768-4777.
Luo, G., Tian, Y., Bychina, M. and Ehlig-Economides, C., 2019. Production-strategy insights using machine learning: application for Bakken Shale. SPE Reservoir Evaluation Engineering, 22(03): 800-816.
Luo, S.-H., Xiao, L.-Z., Jin, Y., Liao, G.-Z., Xu, B.-S., Zhou, J. and Liang, C., 2022. A machine learning framework for low-field NMR data processing. Petroleum Science.
Molnar, C., 2020. Interpretable machine learning. Lulu. com.
Oliver, D.S., Reynolds, A.C. and Liu, N., 2008. Inverse theory for petroleum reservoir characterization and history matching.
Reynolds, A.C., Zafari, M. and Li, G., 2006. Iterative forms of the ensemble Kalman filter, ECMOR X-10th European Conference on the Mathematics of Oil Recovery.
Ribeiro, M.T., Singh, S. and Guestrin, C., 2016. " Why should i trust you?" Explaining the predictions of any classifier, Proceedings of the 22nd ACM SIGKDD international conference on knowledge discovery and data mining, pp. 1135-1144.
Rogers, S.J., Fang, J., Karr, C. and Stanley, D., 1992. Determination of lithology from well logs using a neural network. AAPG bulletin, 76(5): 731-739.
Roth, A.E., 1988. The Shapley value: essays in honor of Lloyd S. Shapley. Cambridge University Press.
Shapley, L.S., 1951. Notes on the N-person Game. Rand Corporation.
Shen, F., Ren, S.S., Zhang, X.Y., Luo, H.W. and Feng, C.M., 2021. A digital twin-based approach for optimization and prediction of oil and gas production. Mathematical Problems in Engineering, 2021.
Shwartz-Ziv, R. and Tishby, N., 2017. Opening the black box of deep neural networks via information. arXiv preprint arXiv:.00810.

Song, X., Liu, Y., Xue, L., Wang, J., Zhang, J., Wang, J., Jiang, L. and Cheng, Z., 2020. Time-series well performance prediction based on Long Short-Term Memory (LSTM) neural network model. Journal of Petroleum Science Engineering, 186: 106682.

Sun, H., Chawathe, A., Hoteit, H., Shi, X. and Li, L., 2015. Understanding shale gas flow behavior using numerical simulation. SPE Journal, 20(01): 142-154.

Tang, J., Fan, B., Xiao, L., Tian, S., Zhang, F., Zhang, L. and Weitz, D., 2021. A new ensemble machine-learning framework for searching sweet spots in shale reservoirs. SPE Journal, 26(01): 482-497.

Wang, S. and Chen, S., 2019. Insights to fracture stimulation design in unconventional reservoirs based on machine learning modeling. Journal of Petroleum Science Engineering, 174: 682-695.

Wang, S., Chen, Z. and Chen, S., 2019. Applicability of deep neural networks on production forecasting in Bakken shale reservoirs. Journal of Petroleum Science Engineering, 179: 112-125.

Yao, J., Sun, H., Fan, D.-y., Wang, C.-c. and Sun, Z.-x., 2013. Numerical simulation of gas transport mechanisms in tight shale gas reservoirs. Petroleum Science, 10(4): 528-537.

Zhang, D., 2001. Stochastic Methods for Flow in Porous Media: Coping with Uncertainties. Elsevier.

Zhang, D., Chen, Y. and Meng, J., 2018. Synthetic well logs generation via Recurrent Neural Networks. Petroleum Exploration Development, 45(4): 629-639.

Zhang, H., Liang, Y., Zhang, W., Wang, B., Yan, X. and Liao, Q., 2017a. A unified MILP model for topological structure of production well gathering pipeline network. Journal of Petroleum Science Engineering, 152: 284-293.

Zhang, H., Liang, Y., Zhou, X., Yan, X., Qian, C. and Liao, Q., 2017b. Sensitivity analysis and optimal operation control for large-scale waterflooding pipeline network of oilfield. Journal of Petroleum Science Engineering, 154: 38-48.

Zhang, Q., Yang, Y., Ma, H. and Wu, Y.N., 2019. Interpreting cnns via decision trees, Proceedings of the IEEE/CVF conference on computer vision and pattern recognition, pp. 6261-6270.

Zhao, X., Zhang, K., Chen, G., Xue, X., Yao, C., Wang, J., Yang, Y., Zhao, H. and Yao, J., 2020. Surrogate-assisted differential evolution for production optimization with nonlinear state constraints. Journal of Petroleum Science Engineering, 194: 107441.